\documentclass[10pt]{article} 
\usepackage[preprint]{tmlr}

\usepackage{amsmath,amsfonts,bm}

\def\eqref#1{equation~\ref{#1}}

\def\1{\bm{1}}

\DeclareMathAlphabet{\mathsfit}{\encodingdefault}{\sfdefault}{m}{sl}
\SetMathAlphabet{\mathsfit}{bold}{\encodingdefault}{\sfdefault}{bx}{n}

\usepackage{hyperref}
\usepackage{url}

\usepackage{booktabs}
\usepackage{multicol}
\usepackage{multirow}
\usepackage[table]{xcolor}
\usepackage{colortbl}
\usepackage{amsmath} 
\usepackage{amsfonts}
\usepackage{xcolor} 
\usepackage{tcolorbox}
\usepackage{latexsym}
\usepackage{amssymb}
\usepackage[whole]{bxcjkjatype}
\usepackage{colortbl}
\usepackage{wrapfig}
\usepackage{wrapstuff}
\usepackage{arydshln}
\usepackage{hyperref}
\usepackage{arydshln}

\title{\textsc{TextTIGER}: Text-based Intelligent Generation with Entity Prompt Refinement for Text-to-Image Generation}

\author{
  Shintaro Ozaki ${}^{\alpha}$ \hspace{5pt}
  Tomoyuki Jinnno ${}^{\alpha}$ \hspace{5pt}
  Kazuki Hayashi ${}^{\alpha}$  \hspace{5pt}
  Yusuke Sakai ${}^{\alpha}$ \hspace{5pt}
  Jingun Kwon ${}^{\gamma}$ \\ [5pt]  
  \textbf{Hidetaka Kamigaito} ${}^{\alpha}$ \hspace{5pt}
  \textbf{Katsuhiko Hayashi} ${}^{\beta}$ \hspace{5pt}
  \textbf{Manabu Okumura} ${}^{\delta}$ \hspace{5pt}
  \textbf{Taro Watanabe} ${}^{\alpha}$  \\  [5pt]
${}^{\alpha}$ Nara Institute of Science and Technology (NAIST), Japan \hspace{7pt} \\
${}^{\beta}$ The University of Tokyo, Japan \\
${}^{\gamma}$ Chungnam National University, Korea\hspace{7pt} 
${}^{\delta}$ Institute of Science Tokyo, Japan
\\ [5pt]
\texttt{\{ozaki.shintaro.ou6, kamigaito.h, taro.watanabe\}@naist.ac.jp} \\ [3pt]
}

\def\month{MM}  
\def\year{YYYY} 
\def\openreview{\url{https://openreview.net/forum?id=XXXX}} 

\begin{document}

\maketitle

\begin{abstract}
When generating images from prompts that include specific entities, the model must retain as much entity-specific knowledge as possible.
However, the number of entities is almost countless, and new entities emerge; memorizing all of them completely is not realistic.
To bridge this gap, our work proposes Text-based Intelligent Generation with Entity Prompt Refinement (\textsc{TextTIGER}). 
\textsc{TextTIGER} strengthens knowledge about entities that appear in the prompt by augmenting external information and then summarizes the expanded descriptions with large language models, preventing performance degradation that arises from excessively long inputs.
To evaluate our method, we construct a new dataset consisting of captions, images, detailed descriptions, and lists of entities.
Experiments with multiple image generation models show that \textsc{TextTIGER} improves image generation performance on widely used evaluation metrics compared with prompts that use captions alone.
In addition, using Multimodal LLM (MLLM)-as-a-judge, which shows a strong correlation with human evaluation, we demonstrate that our method consistently achieves higher scores, which underscores its effectiveness.
These results show that strengthening entity-related descriptions, summarizing them, and refining prompts to an appropriate length leads to substantial improvements in image generation performance.
We will release the created dataset and code upon acceptance.
\end{abstract}

\begin{figure}[t]
    \centering
    \includegraphics[width=0.9\linewidth]{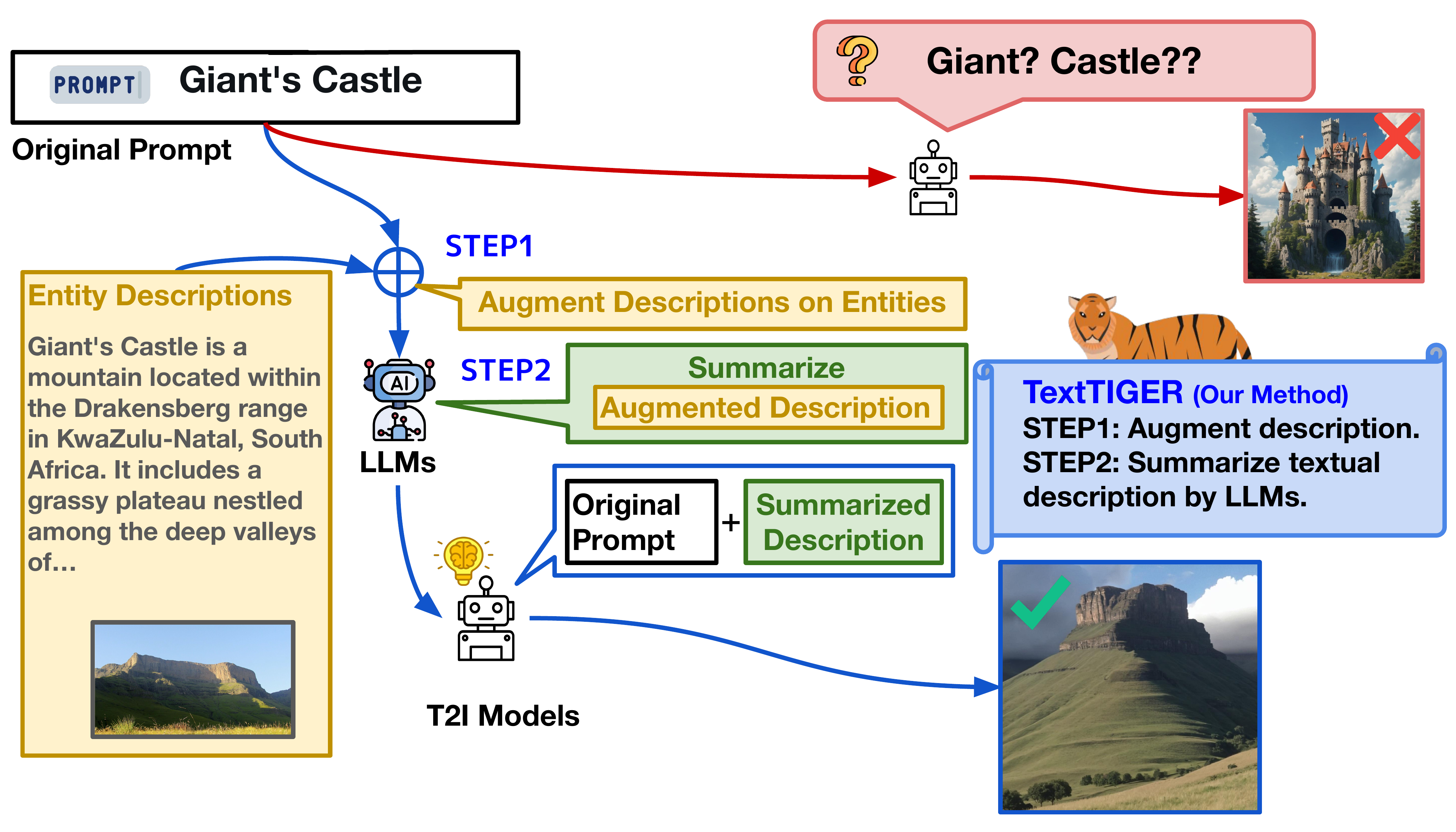}
    \caption{Overview of the proposed method. 
    Our work (1) expands knowledge about entities and (2) summarizes the expanded descriptions to an appropriate length with LLMs, thereby improving the ability of image generation models to handle entities.
    }
    \label{fig:top}
\end{figure}

\section{Introduction}
Text-to-Image generation is a task that generates images from a given text~\citep{zhang2023survey,croitoru2023diffusion} with a wide range of applications, including concept image creation and diagram generation~\citep{zhang2023text}.
To generate images from textual information, image generation models such as Stable Diffusion~\citep{Rombach_2022_CVPR} adopt an architecture that combines a text encoder with a diffusion model~\citep{ho2020denoising}.
These models require carefully designed prompts to reflect the intended image content
~\citep{jeon-etal-2025-iterative,lyu2024imageanythingreasoningcoherenttrainingfree,zhan-etal-2024-prompt, zhang2023survey}.

In this process, the model should retain as much entity-specific knowledge as possible to generate images that meet user expectations.
Such entities include proper nouns in the prompt, such as names of rivers, castles, and mountains
~\citep{seyler-etal-2018-study,yamada-etal-2020-luke,yamada-etal-2018-representation,yamada-etal-2017-learning,gabrilovich2007computing}.\footnote{In our study, we define an entity as a named entity at the proper expression level, referring to a specific instance such as ``Golden Gate Bridge'' rather than an abstract concept such as ``bridge''~\citep{seyler-etal-2018-study,huang2026kitten}.}

However, even large-scale image generation models cannot fully retain such knowledge or continuously acquire the latest information, as it demands substantial costs, i.e., the need to keep crawling for up-to-date information and to continuously train billion-scale models.
Understanding entities correctly plays a crucial role in aligning with user intent in tasks such as an advertisement image generation task~\citep{du2024towards,mita-etal-2024-striking}.
For example, as shown in Figure~\ref{fig:top}, when the prompt ``Giant's Castle'' is given, an image generation model may fail to interpret the entity correctly. 
Here, ``Giant's Castle'' refers to a mountain located in South Africa.\footnote{\url{https://en.wikipedia.org/wiki/Giant\%27s\_Castle}}
Moreover, simply appending externally retrieved information as a long context to the prompt does not enable effective and accurate processing, due to token length constraints such as a 512-token limit of text encoders~\citep{tan2024empirical,zhang2024long}.

To address the limitation in entity understanding, we construct a new dataset  consisting of image-caption pairs with annotated entity mentions that includes detailed descriptions for each entity, enabling systematic evaluation of how adding external knowledge about entities influences image generation quality.
Based on this dataset, we propose a new method called Text-based Intelligent Generation with Entity Prompt Refinement (\textsc{TextTIGER}).
Our method first retrieves entity-specific knowledge from external sources and expands the original prompt.
For example, as shown in Figure~\ref{fig:top}, for the prompt ``Giant's Castle,'' we retrieve additional context such as ``Giant's Castle is a mountain located within the ...,'' which compensates for missing knowledge inside the model.
Second, we leverage large language models (LLMs)~\citep{openai2024gpt4ocard, singh2025openaigpt5card, yang2025qwen3technicalreport, grattafiori2024llama3herdmodels, qwen2025qwen25technicalreport} to summarize the retrieved descriptions concisely.
This step preserves essential information while keeping the prompt within an appropriate token length.
Finally, we generate images from these refined prompts, which mitigates both the model's knowledge limitations and the difficulty of processing long contexts.

Experiments with multiple image generation models together with LLMs on the constructed dataset show that our method substantially outperforms baseline approaches on widely used automatic evaluation metrics.
While performance drops when we simply append descriptions, the performance improves when we summarize them, which supports the importance of concise descriptions of entities.
Furthermore, evaluation results by Multimodal LLM (MLLM)-as-a-judge~\citep{chen2024mllm}, which strongly correlate with human evaluation, indicate that images generated from prompts with entity description summary contain more entity-related content and exhibit greater faithfulness.

\section{Related Work}
\label{related-work}
\paragraph{Vision and Entity Knowledge}
\label{vision-and-entity-knowledge}
In the Vision and Language (V\&L) field, challenges in understanding visual and textual information often reveal the limited generalization ability of V\&L models in generating text from images for applications such as newspapers \citep{lu-etal-2018-entity,liu-etal-2021-visually}, e-commerce \citep{Ma_2022_CVPR}, fashion \citep{rostamzadeh2018fashiongengenerativefashiondataset}, and artworks \citep{Bai_2021_ICCV,hayashi-etal-2024-towards,ozaki-etal-2025-towards}.
Similarly, \citet{kamigaito-etal-2023-table} show that the V\&L model OFA \citep{wang2022ofa} lacks sufficient entity knowledge in image generation tasks.

Several benchmarks also evaluate how well image generation models understand world knowledge~\citep{chen2022reimagenretrievalaugmentedtexttoimagegenerator, zhao2025envisioning,wu2025kris,niu2025wise}.
An extensive study by \citet{chen2022reimagenretrievalaugmentedtexttoimagegenerator} introduced ``EntityDrawBench,'' a dataset covering 250 entities, and pointed out that image generation models lack knowledge about long-tail entities.
\citet{huang2026kitten} introduced the ``KITTEN'' benchmark to evaluate knowledge-intensive generation and found that even the most advanced models often fail to generate entities with accurate visual details.
In experiments across domains such as landmarks, plants, and animals, models including Stable Diffusion~\citep{esser2024scalingrectifiedflowtransformers} and DALL-E 3~\citep{BetkerImprovingIG} produced images with substantial inaccuracies or missing critical features when asked to depict many real-world entities.
These findings indicate that current diffusion models rely heavily on what they learn during training and lack robust factual grounding for many specific entities.

\paragraph{Refinement of Image Generation Prompts}
Researchers have shown that prompt engineering effectively improves image generation~\citep{jeon-etal-2025-iterative,lyu2024imageanythingreasoningcoherenttrainingfree,zhan-etal-2024-prompt}.
\citet{zhan-etal-2024-prompt} refine prompts by training a dedicated text encoder with image representations to generate desired images.
Other work proposes generating images by training models with reinforcement learning~\citep{ghasemi2025comprehensivesurveyreinforcementlearning,schulman2017proximalpolicyoptimizationalgorithms,rafailov2023direct} so that they produce optimized prompts~\citep{hao2023optimizing}.
Methods that refine prompts with external knowledge also exist.
\citet{jeong-etal-2025-culture} point out that models lack cultural understanding and refine prompts with models equipped with external knowledge to produce more appropriate output images.
Image generation approaches based on Retrieval-Augmented Generation~\citep{chen2022reimagenretrievalaugmentedtexttoimagegenerator} also attempt prompt refinement with retrievers, e.g., for abstract prompts~\citep{lyu2024imageanythingreasoningcoherenttrainingfree} and for multiple objects~\citep{yuan-etal-2025-finerag}.

However, although these approaches leverage external knowledge or iterative refinement to improve prompt quality, they do not explicitly focus on supplementing entity-level knowledge.
In particular, prior methods do not retrieve and inject structured, entity-specific factual descriptions to compensate for missing world knowledge in text-to-image models, overlooking the limitation of input length by the text encoder as well.
Instead, they often use external information, such as cultural alignment or safety refinement, without explicitly addressing whether the model has sufficient factual grounding about individual named entities.

\section{Dataset Construction for Entity-Aware Image Generation}
\label{dataset-creation}
To evaluate whether providing rich descriptive information for named entities improves image generation quality, we construct a new dataset.
The existing dataset PopVQA~\citep{haklay-etal-2025-position} provides large-scale image-caption pairs but does not include explicit entity-level information.
As a result, its usefulness is limited in settings where models must understand specific named entities and correctly align them with visual content.
In real-world applications, prompts often include proper nouns and named entities that presuppose background knowledge. Without access to such knowledge, even advanced image generation models may hallucinate incorrect visual content, miss distinctive attributes, or confuse entities.

To address this issue, we extend the original PopVQA by adding background descriptions for all named entities that appear in each caption.
We extend these descriptions through the Wikipedia API.\footnote{\url{https://www.mediawiki.org/wiki/API:Main\_page}}
Specifically, the metadata of PopVQA contains hyperlinks to Wikipedia pages corresponding to entities mentioned in the captions.
We systematically follow these URLs and extract the introductory abstract of each page.
These introductory paragraphs typically provide concise and informative summaries, including the definition, classification, origin, and notable characteristics of the entity.
Such abstracts serve as natural and reliable sources of contextual knowledge, especially for uncommon, ambiguous, or culturally specific entities.
For example, when given the caption ``Liberty at sunset,'' the Wikipedia abstract provides supplementary information such as its \underline{location}, \underline{height}, \underline{appearance}, and \underline{symbolic meaning}.
This knowledge often plays an important role in faithful image generation.

To ensure consistency and quality, we focus on the landmarks and paintings categories and retain only instances for which both the image and the linked Wikipedia page remain accessible at the time of dataset construction.
Under these criteria, we extract 2,764 instances from the landmarks category and 2,245 instances from the paintings category, resulting in 5,009 valid instances in total.
Each instance in our dataset consists of four elements: (1) the original image, (2) the corresponding caption, (3) the retrieved entity descriptions, and (4) the list of entities contained in those descriptions.
The resulting new dataset enables controlled experimental analysis of how access to entity-specific background knowledge influences the behavior of text-to-image generation models. 
Details appear in Appendix~\ref{detailed-dataset-settngs}.

\section{Proposed Method: \textsc{TextTIGER}}
\label{proposed-method}
We propose a method that strengthens entity-specific knowledge by augmenting accurate descriptions of entities that appear in the prompt and then summarizing them to an appropriate length by LLMs, as shown in Figure~\ref{fig:top}.
Our method effectively mitigates two major weaknesses of image generation models, i.e., (1) limited internal knowledge on entities, and (2) difficulty in handling long contexts.
The proposed approach consists of the following 2 steps:
STEP 1. augment entities with informative descriptions (\S~\ref{augmentation}), and STEP 2. summarize the descriptions with an LLM to the appropriate length (\S~\ref{summarization}).

\subsection{STEP 1: Augment Entities with Informative Descriptions}
\label{augmentation}
To enable image generation models to understand entities, we augment externally retrieved, information-rich descriptions to the entities that appear in the prompt.
Specifically, we extract entities contained in the prompt and obtain corresponding descriptions to compensate for the model's limited internal knowledge.

\subsection{STEP 2: Summarize the descriptions using LLMs}
\label{summarization}
For the augmented entity-specific descriptions obtained in STEP 1, we use an LLM to generate summaries that preserve detailed entity information while keeping the length appropriate.
Prior work~\citep{juseon-do-etal-2024-instructcmp} shows that explicitly specifying the input length and the desired number of output tokens helps LLMs manage length constraints effectively. 
From their motivation, we tokenize the augmented descriptions from STEP 1 using the tokenizer of the text encoder in the image generation model and explicitly provide the token count to the LLM, providing the detailed prompts in Appendix~\ref{prompts-for-summarization}.
After this process, we concatenate the summarized entity-specific descriptions to the end of the original caption, forming a new prompt in the format ``(caption + summarized description)'' for image generation.

We refer to this as Text-based Intelligent Generation with Entity Prompt Refinement (\textsc{TextTIGER}).

\section{Experimental Settings}
\subsection{Prompt Formats}
\label{prompt-formats}
To verify whether our proposed method, \textsc{TextTIGER}, properly improves entity-level image generation capability, we additionally compared 4 methods, as introduced in Table~\ref{tab:method-table}.

\begin{enumerate}
\item \underline{\textsc{Cap-Only}}: This setting simply provides the caption that exists in the created dataset to image generation models, validating the baseline performance of image generation models.

\item \underline{\textsc{Aug-Only}}: In this setting, we simply concatenate the entity descriptions to the caption, i.e., applying only the STEP 1 of our approach in \S~\ref{augmentation}.
We define this method to examine whether summarization is necessary.

\item \underline{\textsc{RAG}}: STEP 1 of our proposed method, i.e., \S~\ref{augmentation}, supplements missing knowledge about entities by retrieving external information, which closely relates to the Retrieval-Augmented Generation (RAG)~\citep{lewis2020retrieval} method.  
However, prior work shows that entities are more challenging to handle than general knowledge~\citep{shachar-etal-2025-ner}.  
To evaluate performance under a RAG-based framework, we compare a method that uses BM25~\citep{bm25s} as the retriever, since preliminary experiments show that BM25 achieves the best retrieval performance as demonstrated in Appendix~\ref{app:difference-bw-retriever}.  
As the datastore, we use PopVQA data and additionally collect 109,598 Wikipedia articles related to landmarks and 132,573 articles related to paintings. 
Our method takes the caption as input and concatenates the description retrieved from the datastore to evaluate whether RAG improves performance in this setting.

\item \underline{\textsc{TextTIGER w/o Len}}: STEP 2 of our proposed method (\S~\ref{summarization}) explicitly specifies the token length during summarization to generate prompts that suit image generation.  
To examine whether explicit token control improves prompt quality, we define a variant that performs summarization without specifying the token length.  
This comparison allows us to validate the effectiveness of STEP 2 and to examine whether we can apply prior findings~\citep{juseon-do-etal-2024-instructcmp} to prompt construction.
\end{enumerate}

\begin{table}[t]
    \caption{
    List of experimental settings and comparison methods in our study.
    }
    \small
    \centering
    \resizebox{\linewidth}{!}{
    \begin{tabular}{ll}
    \toprule
    \textbf{Method} & \textbf{Prompt for Image Generation} \\
    \midrule
     \textbf{Cap-Only} & The caption in the dataset.  \\ 
     \textbf{Aug-Only} & The caption + Augmented knowledge from Wikipedia. \\ 
     \textbf{RAG} & The caption + Augmented knowledge retrieved from the datastore. \\
     \textbf{TextTIGER w/o Len} & The caption + Summarized description generated by LLMs. \\ 
     \hdashline
     \textbf{TextTIGER (Our proposed method)} & The caption + Summarized description generated by LLMs with the explicit token length. \\ 
     \bottomrule
    \end{tabular}
    }
    \label{tab:method-table}
\end{table}

\subsection{Summarization Models}
To summarize to an appropriate length, we exploit LLMs with strong summarization capabilities, including Qwen2.5 (72B)~\citep{qwen2025qwen25technicalreport}, Qwen3 (30B)~\citep{yang2025qwen3technicalreport}, and Llama 3.3 (70B)~\citep{grattafiori2024llama3herdmodels} for open models.
As for proprietary models, we use GPT-4o~\citep{openai2024gpt4ocard} and GPT-5~\citep{singh2025openaigpt5card} through API.
Appendix~\ref{detailed-model-settings} provides detailed experimental settings.

\subsection{Image Generation Models}
To account for differences in text encoders, we use five image generation models.
Specifically, we use Dreamlike 2.0 (defined as Dreamlike)~\citep{dreamlike_photoreal_2}, which employs CLIP~\citep{radford2021learning} as its text encoder, PixArt~\citep{chen2023pixartalphafasttrainingdiffusion}, which adopts T5~\citep{raffel2020exploring}, FLUX~\citep{flux2024} and Stable Diffusion 3.5 (SD3.5)~\citep{esser2024scalingrectifiedflowtransformers}, both of which incorporate CLIP and T5, and Qwen-Image (defined as Qwen-Img)~\citep{wu2025qwenimagetechnicalreport}, which uses Qwen~\citep{bai2025qwen25vltechnicalreport} as its encoder.
The maximum token lengths of the text encoders are 77 tokens for CLIP, 512 tokens for T5, and 4,096 tokens for Qwen-Image.

\subsection{Evaluation Metrics}
\label{evaluation-metrics}
To measure how much entity-aware image generation improves, we adopt (1) automatic evaluation metrics, i.e., CLIPScore~\citep{hessel2021clipscore}, DINOScore~\citep{oquab2024dinov2learningrobustvisual}, and PickScore~\citep{kirstain2023pick}, and (2) Multimodal Large Language Models (MLLM)-as-a-judge~\citep{chen2024mllm} evaluated by Vision-Language Models (VLMs)~\citep{qwen2025qwen25technicalreport, microsoft2025phi4minitechnicalreportcompact, gemmateam2025gemma3technicalreport}.

\noindent\textbf{CLIPScore-T:}\,
\label{clipscore-t}
We utilize CLIPScore~\citep{hessel2021clipscore}, which computes the cosine similarity between the hidden states produced by the text encoder and the image encoder of CLIP~\citep{radford2021learning} trained with contrastive learning on image-text pairs, to measure how faithfully an image reflects a given sentence.
We define this metric as CLIPScore-T and use it to measure the similarity between the generated image and the entity used for generation.

\noindent\textbf{CLIPScore-I:}\,
\label{clipscore-i}
Likewise, we compute the similarity between two different images using the hidden states produced by the image encoder of CLIP~\citep{radford2021learning}.
Our work defines this metric as CLIPScore-I and uses it to measure the similarity between the reference image and the generated image.

\noindent\textbf{DINOScore:}\,
\label{dinoscore}
We also compute image-image similarity using the DINO image encoder~\citep{oquab2024dinov2learningrobustvisual}, which they claim to outperform CLIP.
As in CLIPScore-I, we measure the similarity between the reference image and the generated image.
We define this similarity metric based on DINO as DINOScore.

\noindent\textbf{PickScore:}\,
\label{pickscore}
We adopt PickScore~\citep{kirstain2023pick}, which trains on human preference data and estimates the probability that an image aligns with a given textual instruction in a way humans prefer.
PickScore builds on CLIP and, like CLIPScore-T, evaluates the similarity between text and image.

In our work, we use these four automatic metrics to evaluate whether image generation models improve their ability to generate entities accurately.

\begin{table*}[t]
    \centering
    \setlength{\tabcolsep}{2pt}
    \rowcolors{6}{gray!10}{white}
    \caption{Results of experiments on the Landmarks category. We report CLIPScore-T and PickScore as evaluation metrics for text-to-image (Txt-Img) generation, and CLIPScore-I and DINOScore as evaluation metrics for image-to-image (Img-Img) generation.}
    \resizebox{\textwidth}{!}{
    \begin{tabular}{cccccccccccccc}
    \toprule
    \multirow{2.5}{*}{\textbf{T2I Model}}& \multirow{2.5}{*}{\textbf{Cap-Only}}& \multicolumn{2}{c}{\textbf{RAG}} & \multicolumn{5}{c}{\textbf{TextTIGER w/o Len}} & \multicolumn{5}{c}{\textbf{TextTIGER (Proposed Method)}} \\
    \cmidrule(lr){3-4}    \cmidrule(lr){5-9} \cmidrule(lr){10-14}
    &  & \textbf{Aug-Only} & \textbf{BM25} & \textbf{Qwen3} & \textbf{Qwen2.5} & \textbf{Llama3.3} & \textbf{GPT-4o} & \textbf{GPT-5} & \textbf{Qwen3} & \textbf{Qwen2.5} & \textbf{Llama3.3} & \textbf{GPT-4o} & \textbf{GPT-5} \\
    \midrule
\rowcolor{red!10}
\multicolumn{14}{c}{\textbf{CLIPScore-T}} \\
\midrule
Dreamlike & 23.978 & 20.939 & 20.743 & 20.351 & 20.237 & 20.604 & 20.216 & 20.288 & 24.672 & 24.870 & \textbf{25.296} & 24.846 & 24.829 \\
PixArt & 19.106 & 20.293 & 19.630 & 18.955 & 18.962 & 18.986 & 18.960 & 18.953 & 23.106 & 23.108 & \textbf{23.244} & 23.184 & 23.144 \\
FLUX & 21.454 & 20.572 & 20.292 & 19.945 & 19.407 & 19.457 & 19.290 & 19.289 & 23.675 & 23.793 & \textbf{23.990} & 23.686 & 23.669 \\
SD3.5 & 23.882 & 21.432 & 21.258 & 20.571 & 20.753 & 20.840 & 20.799 & 20.746 & 25.032 & 25.397 & \textbf{25.475} & 25.371 & 25.329 \\
Qwen-Img & 22.417 & 17.472 & 17.706 & 20.094 & 19.922 & 19.886 & 19.953 & 19.913 & 24.338 & 24.340 & 24.235 & \textbf{24.445} & 24.414 \\
\midrule
\rowcolor{yellow!10}
\multicolumn{14}{c}{\textbf{CLIPScore-I}} \\
\midrule
Dreamlike & 68.371 & 66.478 & 65.094 & 64.365 & 64.556 & 64.347 & 64.360 & 64.394 & 78.666 & \textbf{79.073} & 78.853 & 78.980 & 78.892 \\
PixArt & 59.679 & 68.515 & 66.432 & 62.484 & 62.729 & 62.338 & 62.590 & 62.593 & 75.864 & 76.252 & 75.435 & \textbf{76.704} & 76.639 \\
FLUX & 66.518 & 69.140 & 67.458 & 64.358 & 64.115 & 63.844 & 63.741 & 63.766 & 77.925 & \textbf{78.433} & 77.988 & 78.356 & 78.250 \\
SD3.5 & 68.990 & 67.450 & 66.043 & 64.762 & 65.414 & 65.424 & 65.304 & 65.336 & 79.334 & \textbf{79.939} & 79.424 & 79.842 & 79.794 \\
Qwen-Img & 69.946 & 47.835 & 47.906 & 64.638 & 64.626 & 64.677 & 64.551 & 64.542 & 79.290 & 79.286 & 78.570 & \textbf{79.672} & 79.563 \\
\midrule
\rowcolor{green!10}
\multicolumn{14}{c}{\textbf{DINOScore}} \\
\midrule
Dreamlike & 29.943 & 27.047 & 23.895 & 33.248 & 33.425 & 34.034 & 33.550 & 33.498 & 45.239 & \textbf{45.557} & 44.202 & 45.245 & 45.105 \\
PixArt & 19.033 & 34.843 & 29.151 & 34.862 & 35.119 & 34.682 & 35.100 & 35.036 & 45.938 & \textbf{46.305} & 43.949 & 45.927 & 45.869 \\
FLUX & 30.015 & 37.438 & 32.629 & 39.924 & 37.743 & 37.568 & 37.704 & 37.645 & 48.972 & \textbf{50.335} & 48.375 & 49.373 & 49.314 \\
SD3.5 & 36.485 & 35.249 & 30.677 & 39.472 & 40.463 & 40.584 & 40.320 & 40.447 & 52.591 & \textbf{53.636} & 51.944 & 53.286 & 53.182 \\
Qwen-Img & 40.417 & 20.242 & 18.198 & 41.460 & 40.935 & 41.365 & 41.416 & 41.439 & 54.716 & 54.029 & 52.200 & \textbf{54.739} & 54.429 \\
\midrule
\rowcolor{blue!10}
\multicolumn{14}{c}{\textbf{PickScore}} \\
\midrule
Dreamlike & 20.556 & 19.092 & 19.934 & 18.517 & 18.556 & 18.517 & 18.584 & 18.586 & 24.756 & 24.784 & 24.743 & 24.821 & \textbf{24.822} \\
PixArt & 20.491 & 19.731 & 20.427 & 18.762 & 18.775 & 18.750 & 18.762 & 18.767 & 25.003 & \textbf{25.018} & 25.016 & 24.947 & 24.939 \\
FLUX & 20.837 & 19.838 & 20.604 & 19.014 & 18.914 & 18.939 & 18.942 & 18.934 & \textbf{25.183} & 25.165 & 25.132 & 25.166 & 25.169 \\
SD3.5 & 20.532 & 19.137 & 19.962 & 18.550 & 18.588 & 18.593 & 18.602 & 18.602 & 24.703 & \textbf{24.747} & 24.658 & 24.731 & 24.729 \\
Qwen-Img & 20.614 & 17.891 & 18.792 & 18.801 & 18.485 & 18.676 & 18.505 & 18.506 & \textbf{24.927} & 24.621 & 24.592 & 24.618 & 24.615 \\
    \bottomrule    
    \end{tabular}
    }
    \label{tab:overall-results-landmarks}
\end{table*}

\noindent\textbf{MLLM-as-a-judge:}\,
\label{mllm-as-a-judge}
\citet{huang2026kitten} proposed ``KITTEN,'' an evaluation framework based on Multimodal LLM (MLLM)-as-a-judge~\citep{chen2024mllm} for evaluating entity-level fidelity in generated images, and reported a high correlation with human evaluation, claiming that it serves as an effective substitute for manual assessment.
In our study, we also adopt KITTEN to evaluate whether our proposed method, \textsc{TextTIGER}, improves entity-aware image generation, and the detailed prompts are provided in Appendix~\ref{prompts-for-kitten}.
Specifically, following KITTEN, we conduct an MLLM-as-a-judge evaluation from two perspectives: \textbf{Entity Alignment} and \textbf{Text Alignment}.
\textbf{Entity Alignment} measures how accurately the generated image reflects the target entity, given a reference image that contains the entity, and VLMs rate this aspect on a 1--5 scale.
\textbf{Text Alignment} measures how faithfully the generated image follows the input text prompt, and VLMs also rate this aspect on a 1--5 scale.
As VLMs that serve as evaluators and support multiple image inputs, we use 3 models: Qwen 2.5~\citep{qwen2025qwen25technicalreport}, Phi 4~\citep{microsoft2025phi4minitechnicalreportcompact}, and Gemma 3~\citep{gemmateam2025gemma3technicalreport}.
In our work, to address the issue that VLMs do not sufficiently understand entities~\citep{hayashi-etal-2024-towards, ozaki-etal-2025-towards,alonso-etal-2025-vision}, we insert a brief description of each entity into the evaluation prompt when applying MLLM-as-a-judge.
Appendix~\ref{detailed-model-settings} provides the details of each model, and Appendix~\ref{prompts-for-kitten} shows the specific prompts we use.

\begin{table*}[t]
    \centering
    \setlength{\tabcolsep}{2pt}
    \rowcolors{6}{gray!10}{white}
    \caption{Experimental results on the Paintings category. 
    The interpretation is the same as in Table~\ref{tab:overall-results-landmarks}.
    }
    \resizebox{\textwidth}{!}{
    \begin{tabular}{cccccccccccccc}
    \toprule
    \multirow{2.5}{*}{\textbf{T2I Model}}& \multirow{2.5}{*}{\textbf{Cap-Only}}& \multicolumn{2}{c}{\textbf{RAG}} & \multicolumn{5}{c}{\textbf{TextTIGER w/o Len}} & \multicolumn{5}{c}{\textbf{TextTIGER (Proposed Method)}} \\
    \cmidrule(lr){3-4}    \cmidrule(lr){5-9} \cmidrule(lr){10-14}
    &  & \textbf{Aug-Only} & \textbf{BM25} & \textbf{Qwen3} & \textbf{Qwen2.5} & \textbf{Llama3.3} & \textbf{GPT-4o} & \textbf{GPT-5} & \textbf{Qwen3} & \textbf{Qwen2.5} & \textbf{Llama3.3} & \textbf{GPT-4o} & \textbf{GPT-5} \\
    \midrule
\rowcolor{red!10}
\multicolumn{14}{c}{\textbf{CLIPScore-T}} \\
\midrule
Dreamlike & 23.830 & 19.833 & 19.708 & 21.959 & 21.832 & 21.530 & 21.337 & 21.572 & 24.414 & 24.610 & \textbf{24.712} & 24.488 & 24.439 \\
PixArt & 21.459 & 20.557 & 19.931 & 21.379 & 21.205 & 21.107 & 20.753 & 20.757 & 23.810 & 23.874 & \textbf{23.930} & 23.659 & 23.696 \\
FLUX& 21.600 & 20.139 & 19.907 & 21.602 & 21.283 & 21.524 & 20.997 & 21.081 & 23.293 & \textbf{23.563} & 23.547 & 23.205 & 23.204 \\
SD3.5 & 23.567 & 20.875 & 20.452 & 21.946 & 21.637 & 22.030 & 21.981 & 22.304 & 24.258 & 24.438 & \textbf{24.599} & 24.128 & 24.135 \\
Qwen-Img & 21.377 & 18.266 & 18.432 & 21.649 & 21.973 & 21.283 & 21.316 & 21.892 & \textbf{24.205} & 23.588 & 23.637 & 23.321 & 23.312 \\
\midrule
\rowcolor{yellow!10}
\multicolumn{14}{c}{\textbf{CLIPScore-I}} \\
\midrule
Dreamlike & 55.861 & 58.167 & 57.851 & 56.194 & 56.654 & 56.309 & 55.951 & 55.933 & \textbf{67.425} & 66.476 & 66.972 & 65.772 & 65.737 \\
PixArt & 56.312 & 59.752 & 58.446 & 57.231 & 56.298 & 56.576 & 56.110 & 55.571 & \textbf{66.714} & 66.062 & 65.896 & 65.383 & 65.455 \\
FLUX& 53.715 & 57.519 & 57.144 & 56.769 & 56.937 & 56.914 & 56.424 & 57.004 & \textbf{63.909} & 62.955 & 62.841 & 62.008 & 61.950 \\
SD3.5 & 56.427 & 57.525 & 57.789 & 58.186 & 57.514 & 57.680 & 58.191 & 57.424 & \textbf{66.252} & 65.555 & 65.769 & 64.524 & 64.471 \\
Qwen-Img & 56.199 & 49.763 & 51.657 & 56.156 & 56.092 & 56.729 & 55.215 & 56.049 & \textbf{68.857} & 67.618 & 66.910 & 67.120 & 67.151 \\
\midrule
\rowcolor{green!10}
\multicolumn{14}{c}{\textbf{DINOScore}} \\
\midrule
Dreamlike & 30.673 & 27.454 & 26.636 & 32.047 & 34.949 & 36.700 & 34.040 & 32.965 & \textbf{44.316} & 42.476 & 43.615 & 41.010 & 40.732 \\
PixArt & 30.622 & 39.980 & 37.742 & 35.487 & 34.764 & 37.385 & 34.540 & 35.312 & \textbf{44.037} & 42.786 & 43.624 & 41.333 & 41.298 \\
FLUX& 24.118 & 32.084 & 30.600 & 36.908 & \textbf{38.218} & 36.275 & 36.390 & 36.412 & 33.839 & 33.459 & 34.046 & 31.523 & 31.335 \\
SD3.5 & 30.465 & 32.520 & 32.224 & 38.880 & 39.296 & 39.074 & 37.479 & 38.324 & 39.366 & 38.307 & \textbf{40.362} & 35.848 & 35.845 \\
Qwen-Img & 33.026 & 24.967 & 26.042 & 39.765 & 39.409 & 40.556 & 38.342 & 39.335 & \textbf{46.228} & 46.167 & 45.632 & 45.072 & 45.345 \\
\midrule
\rowcolor{blue!10}
\multicolumn{14}{c}{\textbf{PickScore}} \\
\midrule
Dreamlike & 20.198 & 18.542 & 19.331 & 18.953 & 19.035 & 19.027 & 19.029 & 19.038 & 24.340 & 24.326 & 24.281 & \textbf{24.367} & 24.361 \\
PixArt & 20.678 & 19.278 & 19.906 & 19.351 & 19.326 & 19.375 & 19.283 & 19.249 & 24.733 & \textbf{24.831} & 24.767 & 24.812 & 24.811 \\
FLUX& 20.863 & 19.541 & 20.167 & 19.512 & 19.656 & 19.595 & 19.550 & 19.556 & 24.850 & 24.913 & 24.890 & \textbf{24.921} & 24.919 \\
SD3.5 & 20.362 & 18.863 & 19.647 & 19.199 & 19.207 & 19.033 & 19.119 & 19.164 & 24.484 & \textbf{24.508} & 24.452 & 24.505 & 24.485 \\
Qwen-Img & 20.278 & 18.105 & 18.967 & 19.401 & 19.249 & 19.381 & 18.740 & 19.091 & \textbf{24.582} & 23.951 & 24.001 & 23.893 & 23.898 \\
    \bottomrule    
    \end{tabular}
    }
    \label{tab:overall-results-paintings}
\end{table*}

\section{Results and Discussions}
\label{results-and-discussions}
Tables~\ref{tab:overall-results-landmarks} and~\ref{tab:overall-results-paintings} present the results of the automatic evaluation metrics as described in \S~\ref{evaluation-metrics}.
In general, focusing on the Landmarks category in Table~\ref{tab:overall-results-landmarks}, \textsc{TextTIGER} achieves the best performance across the metrics, as indicated by the \textbf{bold} scores in the table, showing that augmenting entity-specific content and summarizing it to an appropriate length effectively improves entity-aware image generation.

In contrast, the \textsc{Aug-Only} setting shows little improvement over \textsc{Cap-Only}, indicating that simply appending additional knowledge does not lead to better performance.
Moreover, when we compare \textsc{TextTIGER w/o Len} with the \textsc{TextTIGER}, \textsc{TextTIGER} consistently performs better, supporting the importance of explicitly controlling the summary length to construct prompts that better suit image generation models.
We observe the same trend in the Paintings category shown in Table~\ref{tab:overall-results-paintings}, and these consistent improvements across both domains demonstrate the generalization ability of our method.

\subsection{Does TextTIGER Perform Effectively?}
We showed that \textsc{TextTIGER} shows clear improvements when we compare the baseline.
Taking a closer look at the result, for CLIPScore-T, PixArt improves from 21.459 to 23.930 with summaries generated by Llama 3.3, and Qwen-Img improves from 21.377 to 24.205 with summaries generated by Qwen 3.
For DINOScore, Qwen-Img improves from 30.465 under \textsc{Cap-Only} to 46.228 when we use the optimized prompt summarized by Qwen 3.

Next, we examine whether simply expanding entity knowledge enables image generation models to understand entities. When we compare \textsc{RAG} setting with \textsc{TextTIGER}, \textsc{TextTIGER} consistently achieves better overall performance.
For example, in CLIPScore-I, FLUX achieves 57.525 under the RAG-style expansion, whereas LLM-based summarization raises the score up to 68.857, yielding an improvement of more than 10 points.
Across other automatic evaluation metrics, \textsc{TextTIGER} also outperforms \textsc{Cap-Only}, which demonstrates that \textsc{TextTIGER} effectively enhances entity-aware image generation.

We also analyze whether explicit length control during summarization is necessary.
When we compare \textsc{TextTIGER w/o Len} with \textsc{TextTIGER}, the \textsc{TextTIGER} consistently achieves better performance.
Moreover, \textsc{TextTIGER w/o Len} performs at a level comparable to \textsc{Cap-Only}, which highlights that instructing LLMs to summarize without explicit token-length control can even degrade performance.
In contrast, \textsc{TextTIGER} improves performance, proving that image generation models remain highly sensitive to token length, yet can substantially benefit when the refined prompt fits within an appropriate input length.

Table~\ref{tab:token_length_by_category} reports the measured prompt token lengths for three knowledge expansion methods: \textsc{Aug-Only}, \textsc{TextTIGER w/o Len}, and \textsc{TextTIGER}.
We compute the token counts using the tokenizer of each summarization model.
For the proprietary GPT-4o/5 models, we use the \texttt{tiktoken} library (\url{https://github.com/openai/tiktoken}).
The results in the table further confirm that explicitly controlling the token length during summarization, as proposed by \citet{juseon-do-etal-2024-instructcmp}, improves the performance.

\begin{table}[t]
\small
\centering
\caption{
    Token length statistics ($\text{mean}_{\pm \text{std}}$).
    \textbf{Bold} indicates the results of \textsc{TextTIGER}.
}
\begin{tabular}{c ccc ccc}
\toprule
\multirow{2}{*}{\textbf{Model}} & \multicolumn{3}{c}{\textbf{Landmarks}} & \multicolumn{3}{c}{\textbf{Paintings}} \\
\cmidrule(lr){2-4}\cmidrule(lr){5-7}
 & \textbf{Aug-Only} & \textbf{w/o Len} & \textbf{TextTIGER} & \textbf{Aug-Only} & \textbf{w/o Len} & \textbf{TextTIGER} \\
\midrule
Qwen3 & $733.2_{\pm 408.1}$ & $190.2_{\pm 18.1}$ & $\textbf{75.4}_{\pm 15.1}$ & $467.5_{\pm 384.0}$ & $186.3_{\pm 24.2}$ & $\textbf{74.8}_{\pm 20.1}$ \\
Llama 3.3 & $707.5_{\pm 393.9}$ & $96.9_{\pm 22.8}$ & $\textbf{33.9}_{\pm 9.0}$ & $454.9_{\pm 373.7}$ & $88.8_{\pm 27.9}$ & $\textbf{31.1}_{\pm 9.1}$ \\
Qwen 2.5 & $733.2_{\pm 408.1}$ & $103.9_{\pm 20.4}$ & $\textbf{40.1}_{\pm 7.5}$ & $467.5_{\pm 384.0}$ & $100.5_{\pm 23.4}$ & $\textbf{39.2}_{\pm 8.9}$ \\
GPT-4o/5 & $693.6_{\pm 387.3}$ & $105.2_{\pm 14.1}$ & $\textbf{45.9}_{\pm 5.5}$ & $446.0_{\pm 368.3}$ & $112.4_{\pm 20.2}$ & $\textbf{45.4}_{\pm 5.7}$ \\
\bottomrule
\end{tabular}
\label{tab:token_length_by_category}
\end{table}

\begin{figure}[t]
    \centering
    \includegraphics[width=\linewidth]{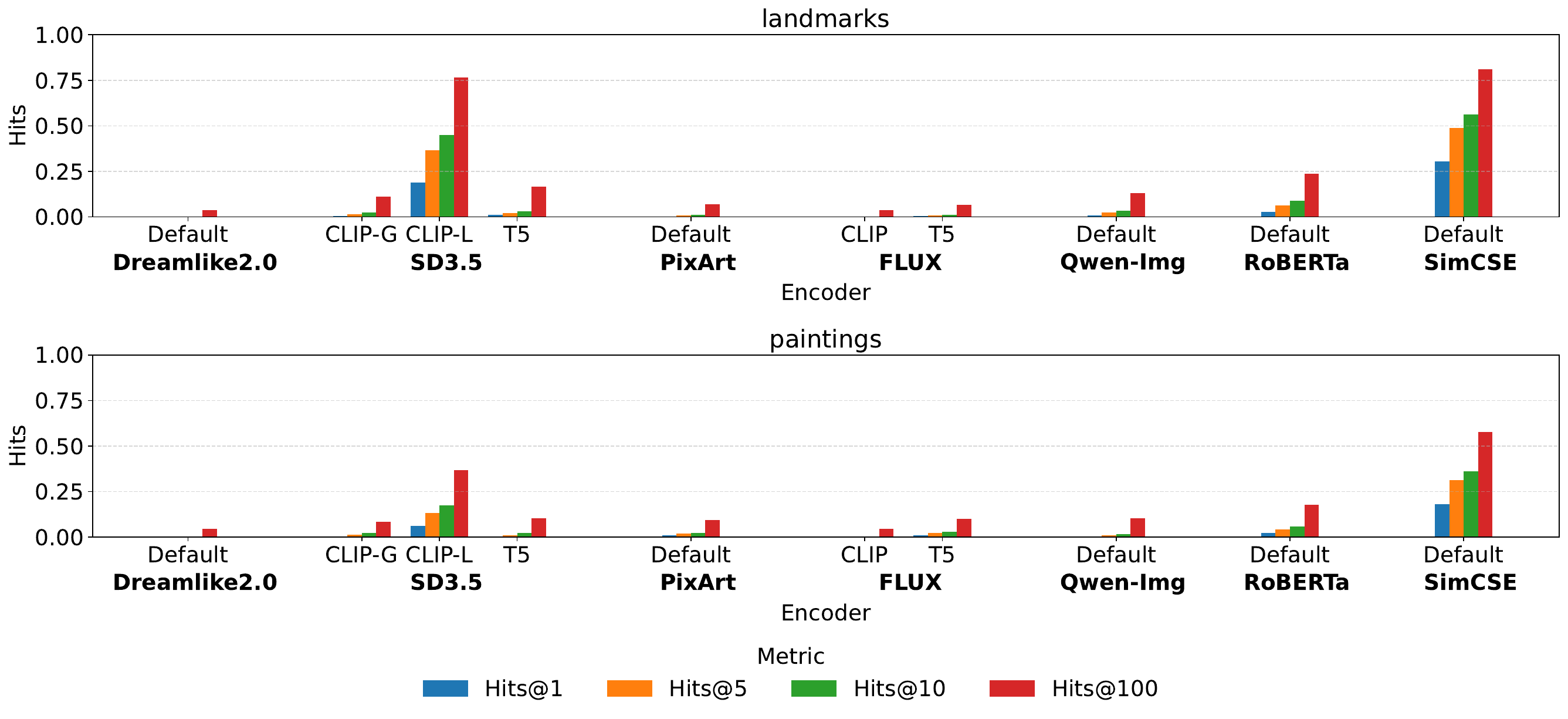}
    \caption{Visualization of how well the text encoders of image generation models understand entities. We examine whether each model can retrieve the correct description when given an entity as input.}
    \label{fig:hits_barplot_models_encoders}
\end{figure}

\subsection{Do Encoders Understand Entities?}
\label{do-encoders-understand-entities?}
Our work showed that supplementing entity information with external knowledge improves image generation performance.
We now ask how well the text encoders of image generation models internally represent entities themselves.
To answer this question, we adopt a knowledge probing framework~\citep{kalo2022kamel,wiland-etal-2024-bear,petroni-etal-2019-language,youssef-etal-2023-give,jinno-etal-2026-cosine}. 
Our work constructs an entity retrieval task using the hidden states produced by each text encoder and evaluates performance with Hits@$k$ ($k$ $\in$ \{1, 5, 10, 100\} in our work).
Specifically, we input a description of each entity and use its embedding representation to retrieve the corresponding gold entity from a datastore.
For the datastore, we take the 5,009 descriptions constructed in \S~\ref{dataset-creation} and rewrite them into ambiguous expressions using GPT-5 so that they do not directly contain the gold entity names,  guaranteeing semantic entity understanding evaluation.

Figure~\ref{fig:hits_barplot_models_encoders} presents the results.
For the Landmarks category (top row), the default encoders of Dreamlike, PixArt, FLUX, and Qwen-Img achieve almost zero Hits@1 and Hits@5, and even Hits@100 remains extremely low.
When we use CLIP-G and CLIP-L text encoders in SD3.5, performance improves to some extent.
However, even then, Hits@100 stays around 0.7, and accuracy at smaller $k$ remains low.
This observation is the same trend in the Paintings category (bottom row).
The text encoders of image generation models fail to reliably identify the correct entity from ambiguous descriptions.
In contrast, the sentence embedding models introduced for comparison, SimCSE~\citep{gao-etal-2021-simcse} and RoBERTa~\citep{liu2019robertarobustlyoptimizedbert}, achieve much higher performance.
In Landmarks, SimCSE exceeds 0.8 in Hits@100 and clearly outperforms image generation encoders in Hits@10 and Hits@5 as well.
We observe the same pattern in Paintings, where SimCSE consistently shows strong retrieval performance, highlighting the difference between text encoders trained for NLP tasks and those used in image generation models.
These findings suggest that although text encoders in image generation models suffice for image conditioning, they struggle with knowledge-intensive reasoning that requires uniquely identifying entities from ambiguous descriptions.
In other words, their internal entity knowledge remains limited, supporting our motivation that supplementing entity knowledge with external information effectively addresses this limitation.

\begin{figure}[t]
    \centering
    \includegraphics[width=\linewidth]{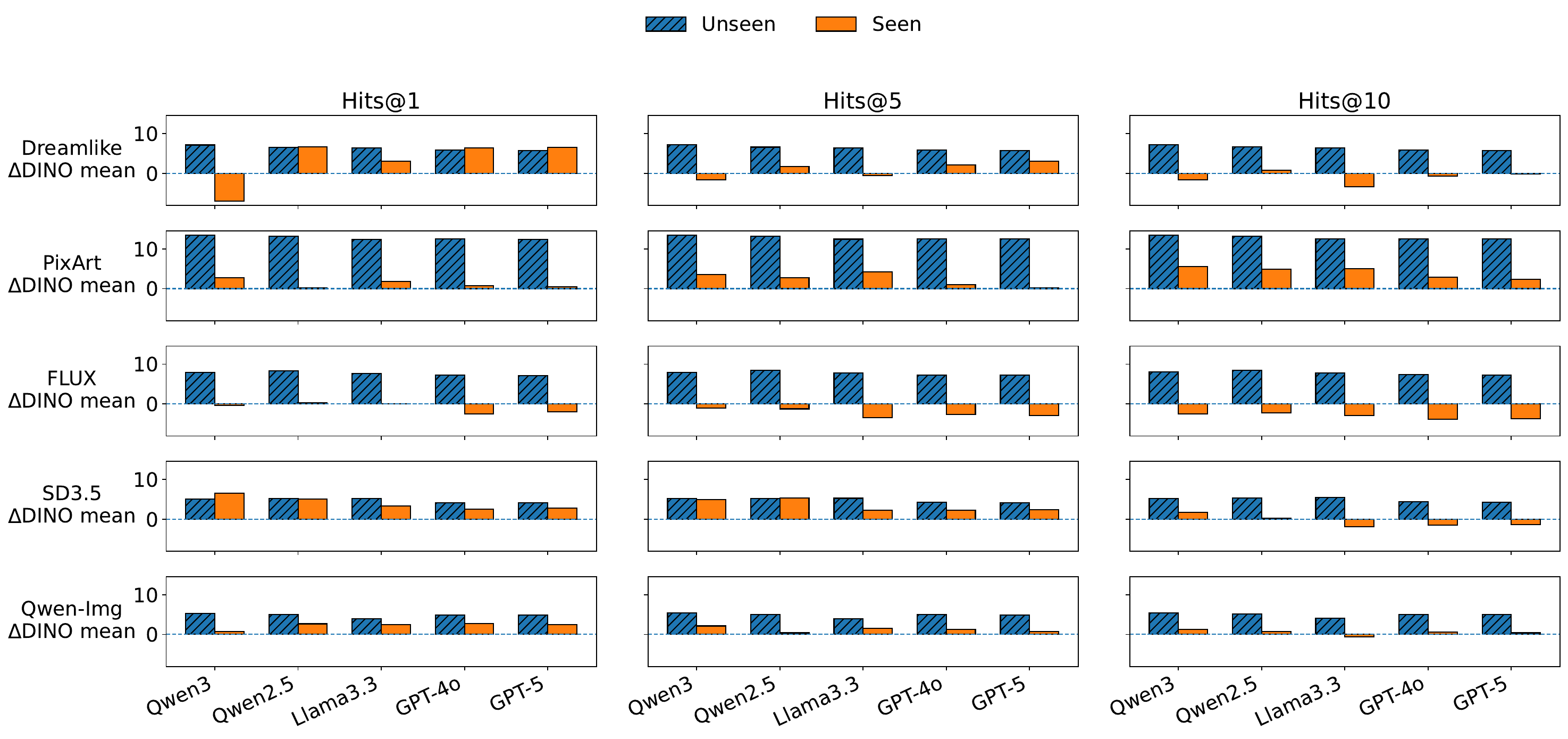}
    \caption{
    Visualization of whether our proposed method, \textsc{TextTIGER}, effectively handles unseen entities. 
    The $y$-axis shows the difference in DINOScore between \textsc{Cap-Only} and \textsc{TextTIGER}.
    The blue bins represent the subset of previously unseen entities, and the orange bins represent the subset of seen entities.
    }
    \label{fig:dino_hits}
\end{figure}

\subsection{Do Descriptions Support Unseen Entities?}
To verify whether our augmented approach, as introduced in \S~\ref{augmentation}, truly benefits ``unseen'' entities, we split entities into two groups based on the retrieval results in \S~\ref{do-encoders-understand-entities?} under the \textsc{Cap-Only} setting, i.e., ``seen'' (Hits=1) and ``unseen'' (Hits=0).
We then computed the difference in DINOScore as introduced in \S~\ref{dinoscore} between \textsc{Cap-Only} and \textsc{TextTIGER} as $\Delta$DINO and analyzed how the effect varies depending on whether the entity was known or unknown.
Figure~\ref{fig:dino_hits} presents the results.
\footnote{Among the automatic evaluation metrics, we base our analysis on DINOScore in our study.We choose DINOScore because it shows the widest score range in Tables~\ref{tab:overall-results-landmarks} and~\ref{tab:overall-results-paintings}, which facilitates clearer analysis.}
The most notable observation is that for Hits=0, namely cases where the text encoder failed to correctly identify the entity (blue bars), $\Delta$DINO consistently shows positive values across many models.
In particular, PixArt and FLUX exhibit large positive improvements for Hits@1, Hits@5, and Hits@10, indicating that adding external descriptions markedly improves visual alignment when the model lacks sufficient internal entity knowledge.
This finding demonstrates that even when the model does not internally encode adequate entity knowledge, structured external information can improve generation quality.

In contrast, for Hits=1, where the encoder already retrieved the correct entity (orange bars), $\Delta$DINO does not consistently remain positive.
Depending on the model and the value of $k$, the improvement becomes small or slightly negative.
For example, FLUX shows negative differences in some Hits=1 cases, suggesting that when the model already encodes the entity to some extent, additional descriptions do not necessarily provide further benefits and may even introduce redundancy.
We observe similar patterns in SD3.5 and Qwen-Img. 
While unknown entities consistently benefit from augmentation, the effect on known entities depends on the model, indicating that external descriptions mainly compensate for missing internal representations rather than further strengthening already encoded knowledge.

\begin{table}[t]
    \caption{
    Results of the MLLM-as-a-judge-based KITTEN text alignment evaluation (Txt-Img).
    C denotes the \textsc{Cap-Only} (Baseline) method. Q3, Q25, L3, G4, and G5 represent image generation results using summaries produced by Qwen3, Qwen2.5, Llama3.3, GPT-4o, and GPT-5, respectively.
    }
    \centering
    \setlength{\tabcolsep}{2.5pt}
    \resizebox{\linewidth}{!}{
    \begin{tabular}{
    c 
    >{\columncolor{gray!10}}c c c c c c|
    >{\columncolor{gray!10}}c c c c c c|
    >{\columncolor{gray!10}}c c c c c c
    }    
    \toprule
    \multirow{3.75}{*}{\textbf{T2I Model}}  & \multicolumn{18}{c}{\textbf{Evaluator}} \\
    \cmidrule(lr){2-19}
    & \multicolumn{6}{c}{\textbf{Gemma3}} & \multicolumn{6}{c}{\textbf{Qwen2.5 VL}} & \multicolumn{6}{c}{\textbf{Phi4}} \\
    \cmidrule(lr){2-7} \cmidrule(lr){8-13} \cmidrule(lr){14-19} 
    & \textbf{C} & \textbf{Q3} & \textbf{Q25} & \textbf{L3} & \textbf{G4} & \textbf{G5}
    & \textbf{C} & \textbf{Q3} & \textbf{Q25} & \textbf{L3} & \textbf{G4} & \textbf{G5}
    & \textbf{C} & \textbf{Q3} & \textbf{Q25} & \textbf{L3} & \textbf{G4} & \textbf{G5} \\
    \midrule
Dreamlike & 3.38 & 4.40 & 4.39 & 4.42 & 4.38 & 4.39 & 2.62 & 3.63 & 3.64 & 3.64 & 3.65 & 3.64 & 2.59 & 3.55 & 3.57 & 3.57 & 3.59 & 3.58 \\
PixArt & 3.11 & 4.48 & 4.49 & 4.50 & 4.46 & 4.48 & 2.51 & 3.72 & 3.73 & 3.74 & 3.72 & 3.72 & 2.52 & 3.70 & 3.75 & 3.78 & 3.74 & 3.75 \\
FLUX & 3.24 & 4.38 & 4.40 & 4.39 & 4.35 & 4.36 & 2.37 & 3.61 & 3.62 & 3.61 & 3.60 & 3.59 & 2.58 & 3.63 & 3.67 & 3.65 & 3.63 & 3.64 \\
SD 3.5 & 3.44 & 4.44 & 4.46 & 4.45 & 4.43 & 4.42 & 2.57 & 3.71 & 3.72 & 3.69 & 3.71 & 3.71 & 2.54 & 3.66 & 3.69 & 3.69 & 3.67 & 3.66 \\
Qwen-Img & 3.43 & 4.47 & 4.33 & 4.30 & 4.28 & 4.28 & 2.53 & 3.70 & 3.59 & 3.52 & 3.57 & 3.57 & 2.65 & 3.56 & 3.25 & 3.24 & 3.24 & 3.23 \\
\bottomrule
    \end{tabular}
    }
    \label{tab:kitten_text_eval}
\end{table}

\begin{table}[t]
    \caption{
    Results of the KITTEN entity alignment evaluation (Img-Img).
    See Table~\ref{tab:kitten_text_eval} for details.
    }
    \centering
    \setlength{\tabcolsep}{2.5pt}
    \resizebox{\linewidth}{!}{
    \begin{tabular}{
    c 
    >{\columncolor{gray!10}}c c c c c c|
    >{\columncolor{gray!10}}c c c c c c|
    >{\columncolor{gray!10}}c c c c c c
    }    
    \toprule
    \multirow{3.75}{*}{\textbf{T2I Model}}  & \multicolumn{18}{c}{\textbf{Evaluator}} \\
    \cmidrule(lr){2-19}
    & \multicolumn{6}{c}{\textbf{Gemma3}} & \multicolumn{6}{c}{\textbf{Qwen2.5 VL}} & \multicolumn{6}{c}{\textbf{Phi4}} \\
    \cmidrule(lr){2-7} \cmidrule(lr){8-13} \cmidrule(lr){14-19} 
    & \textbf{C} & \textbf{Q3} & \textbf{Q25} & \textbf{L3} & \textbf{G4} & \textbf{G5}
    & \textbf{C} & \textbf{Q3} & \textbf{Q25} & \textbf{L3} & \textbf{G4} & \textbf{G5}
    & \textbf{C} & \textbf{Q3} & \textbf{Q25} & \textbf{L3} & \textbf{G4} & \textbf{G5} \\    
    \midrule
Dreamlike & 2.44 & 3.47 & 3.50 & 3.47 & 3.47 & 3.47 & 1.43 & 2.47 & 2.48 & 2.45 & 2.43 & 2.45 & 2.03 & 3.09 & 3.11 & 3.07 & 3.07 & 3.06 \\
PixArt & 2.22 & 3.62 & 3.64 & 3.60 & 3.61 & 3.61 & 1.18 & 2.64 & 2.69 & 2.64 & 2.63 & 2.62 & 1.73 & 3.22 & 3.27 & 3.27 & 3.21 & 3.23 \\
FLUX & 2.35 & 3.54 & 3.58 & 3.53 & 3.55 & 3.56 & 1.28 & 2.53 & 2.60 & 2.50 & 2.51 & 2.53 & 2.05 & 3.15 & 3.23 & 3.17 & 3.14 & 3.13 \\
SD3.5 & 2.59 & 3.64 & 3.68 & 3.64 & 3.66 & 3.65 & 1.67 & 2.76 & 2.85 & 2.79 & 2.81 & 2.81 & 2.28 & 3.30 & 3.35 & 3.33 & 3.28 & 3.29 \\
Qwen-Img & 2.60 & 3.71 & 3.62 & 3.54 & 3.63 & 3.63 & 1.62 & 2.78 & 2.71 & 2.59 & 2.70 & 2.70 & 2.34 & 3.27 & 3.14 & 3.07 & 3.11 & 3.10 \\

\bottomrule
    \end{tabular}
    }
    \label{tab:kitten_entity_eval}
\end{table}



\begin{table}[t]
    \centering
    \caption{Actual images generated by SD 3.5 alongside the reference images and their evaluation scores.
    The scores are reported in the order of \{DINO / Gemma3 / Qwen2.5 / Phi4\}.
    }
    \setlength{\tabcolsep}{6pt}
    \resizebox{\linewidth}{!}{
    \begin{tabular}{c cccc}
    \toprule
    \multirow{2.5}{*}{\textbf{Pattern}} & \multicolumn{2}{c}{\textbf{Landmarks}} & \multicolumn{2}{c}{\textbf{Paintings}} \\
    \cmidrule(lr){2-3} \cmidrule(lr){4-5}
    & \textbf{Białowieża Forest} & \textbf{Po-i-Kalyan} & \textbf{Sacred conversation} & \textbf{Solly Madonna} \\
    \cmidrule(lr){1-5}
    \multirow{-4}{*}{Ref. Img} & 
    \includegraphics[width=1.5cm]{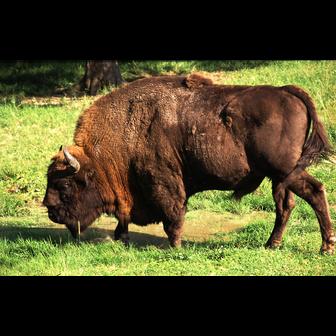} &
    \includegraphics[width=1.5cm]{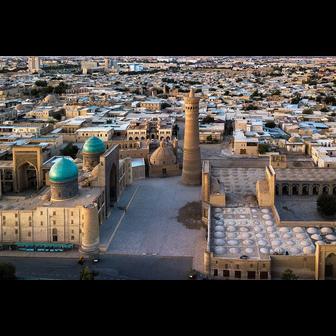} & 
    \includegraphics[width=1.5cm]{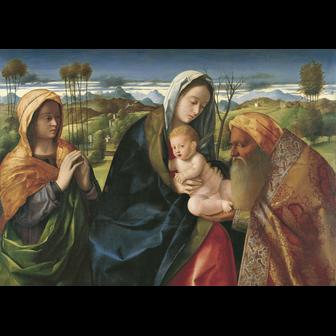} &
    \includegraphics[width=1.5cm]{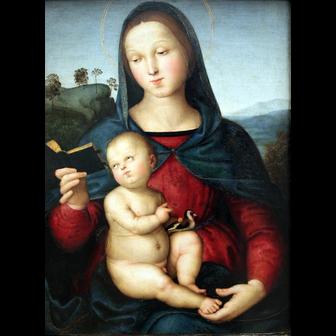} \\
    \cmidrule(lr){2-5}
    \multirow{-3}{*}{Cap-Only} & 
    \includegraphics[width=1.5cm]{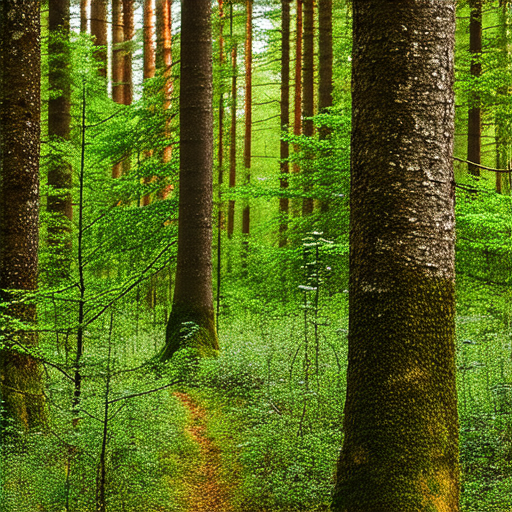} & \includegraphics[width=1.5cm]{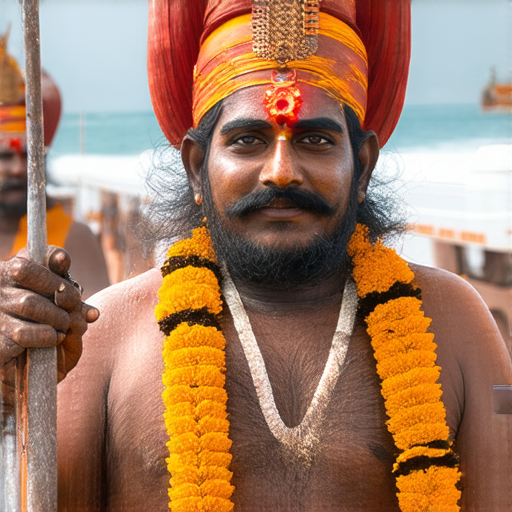} & \includegraphics[width=1.5cm]{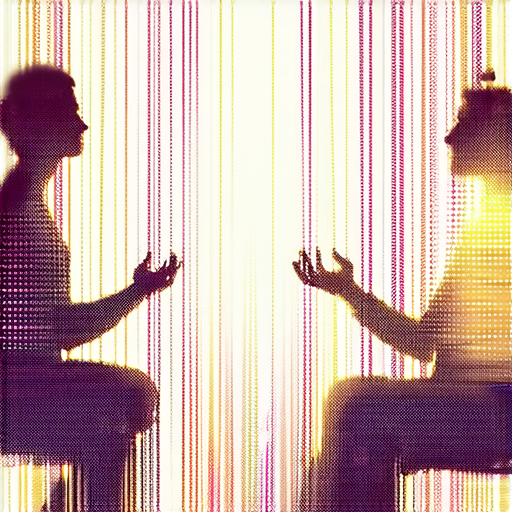} & \includegraphics[width=1.5cm]{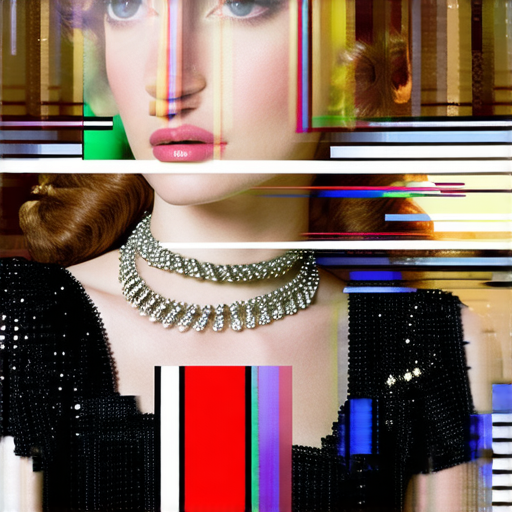} \\
    & 3.12 / 2 / 2/ 1 & 2.07 / 1 / 1 / 1 & 3.25 / 1 / 1 / 1 & 3.18 / 2 / 1 / 1 \\
    \midrule
    \multicolumn{5}{l}{\textbf{Proposed Method (TextTIGER)}} \\
    \midrule
    \multirow{-3}{*}{Qwen3} & 
    \includegraphics[width=1.5cm]{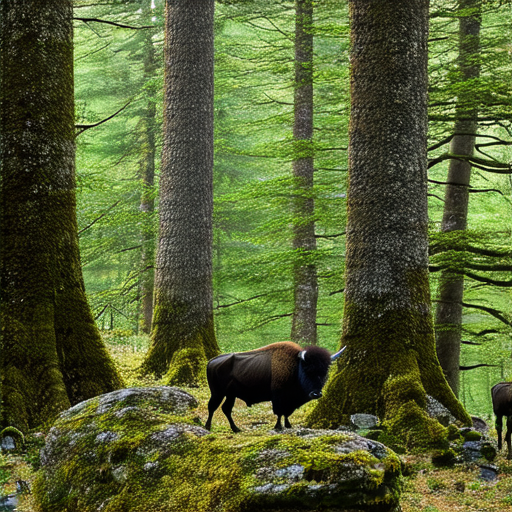} & 
    \includegraphics[width=1.5cm]{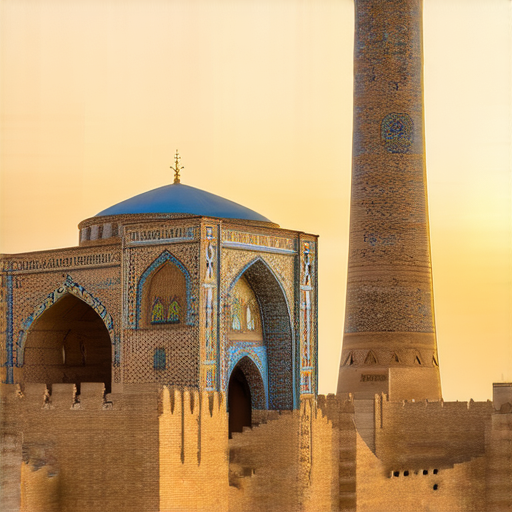} & 
    \includegraphics[width=1.5cm]{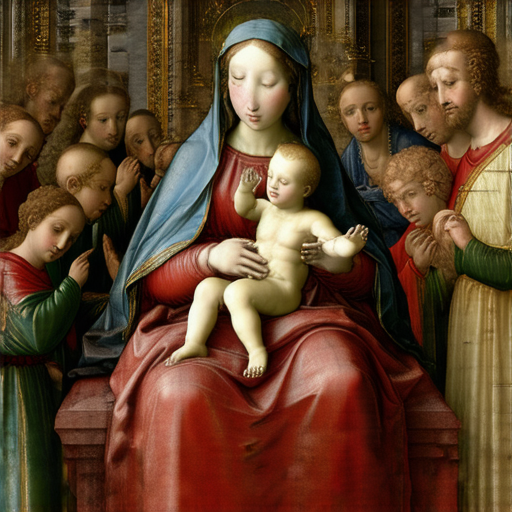} & 
    \includegraphics[width=1.5cm]{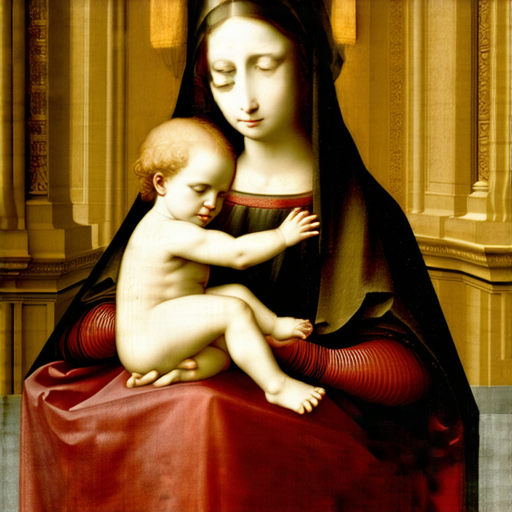} \\
    & 85.12 / 4 / 4 / 3 & 80.25 / 4 / 5 / 3 & 85.75 / 4 / 4 / 3 & 84.50 / 4 / 4 / 4 \\
    \cmidrule(lr){2-5}
    \multirow{-3}{*}{Llama3.3} & 
    \includegraphics[width=1.5cm]{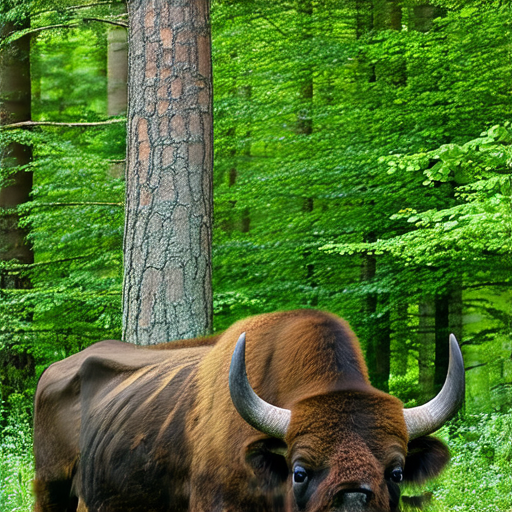} & 
    \includegraphics[width=1.5cm]{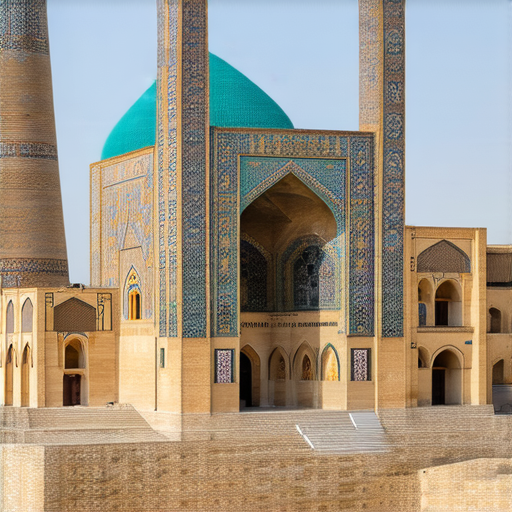} & 
    \includegraphics[width=1.5cm]{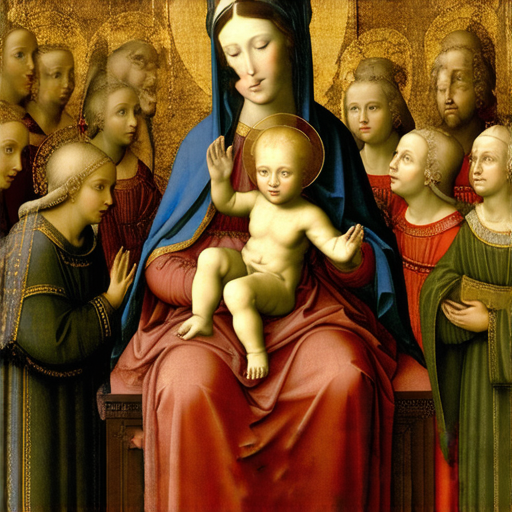} & 
    \includegraphics[width=1.5cm]{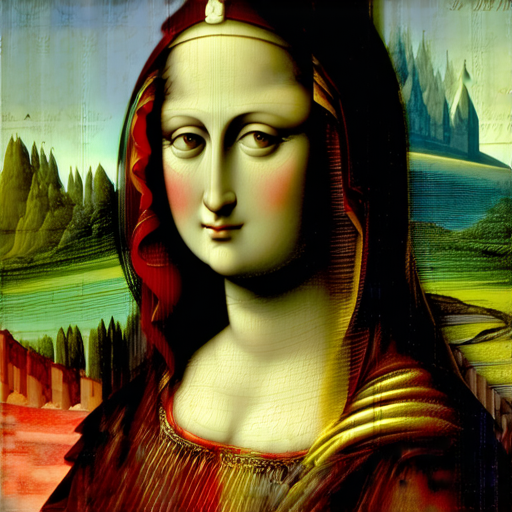} \\
    & 90.21 / 4 / 4 / 4 & 82.23 / 4 / 5 / 3 & 78.234 / 4 / 4 / 3 & 52.12 / 1 / 2 / 2 \\
    \bottomrule
    \end{tabular}
    }
    \label{tab:actual_samepls_generated_by_sd3.5}
\end{table}

\subsection{Results of the MLLM-as-a-judge}
The results of MLLM-as-a-judge shown in Tables~\ref{tab:kitten_entity_eval} and~\ref{tab:kitten_text_eval} demonstrate that introducing external knowledge clearly improves the entity-related capabilities of image generation models in our study.
When we look at entity alignment (Img-Img), compared with \textsc{Cap-Only}, all settings that incorporate external knowledge through summaries generated by Qwen3, Qwen2.5, Llama3.3, GPT-4o, and GPT-5 consistently achieve higher scores across all image generation models.
For example, under the Gemma3 evaluator, Dreamlike improves from 2.44 in \textsc{Cap-Only} to approximately 3.47. PixArt increases from 2.22 to above 3.6, and SD3.5 rises from 2.59 to the high 3.6 range.
We observe similar trends with Qwen2.5 VL and Phi4 as evaluators.
Models that remain in the low 1--2 range under \textsc{Cap-Only} consistently improve to around 2.5--3.3 after adding external knowledge, indicating that generated images reflect entity-specific characteristics more accurately when compared with reference images.
The externally augmented entity descriptions contribute directly to improved visual alignment.

We observe similar improvements in text alignment (Txt-Img).
With the Gemma3 evaluator, Dreamlike increases from 3.38 under \textsc{Cap-Only} to around 4.4, and PixArt improves from 3.11 to approximately 4.48.
FLUX and SD3.5 also gain nearly one full point compared with \textsc{Cap-Only}.
These gains indicate that prompts enriched with external knowledge strengthen the consistency between textual instructions and generated images.
Under Qwen2.5 VL and Phi4 evaluators, \textsc{Cap-Only} remains around 2.3--2.6, whereas knowledge-augmented settings rise to approximately 3.6--3.7, showing that models reflect entity information in the prompt more faithfully after augmentation.
Importantly, these improvements do not depend on a specific image generation model or evaluator.
Dreamlike, PixArt, FLUX, SD3.5, and Qwen-Img all outperform \textsc{Cap-Only} in both entity alignment and text alignment.
These consistent gains suggest that augmenting entity descriptions with external knowledge systematically compensates for internal knowledge gaps.

\begin{table}[t]
    \centering
    \caption{
    Actual generated images produced using summaries by Qwen3, along with the corresponding reference images and their evaluation scores.
    See Table~\ref{tab:actual_samepls_generated_by_sd3.5} for details.
    }
    \centering
    \resizebox{\linewidth}{!}{
    \begin{tabular}{c cccc}
    \toprule
    \multirow{2.5}{*}{\textbf{Pattern}} & \multicolumn{2}{c}{\textbf{Landmarks}} & \multicolumn{2}{c}{\textbf{Paintings}} \\
    \cmidrule(lr){2-3} \cmidrule(lr){4-5}
    & \textbf{Ca' Vendramin Calergi} & \textbf{Palazzo Grassi} & \textbf{Freedom from Want} & \textbf{Palazzo Dario} \\
    \cmidrule(lr){1-5}
    \multirow{-5}{*}{Ref. Img} &
    \includegraphics[width=2cm]{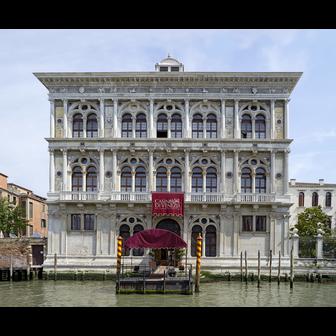} &
    \includegraphics[width=2cm]{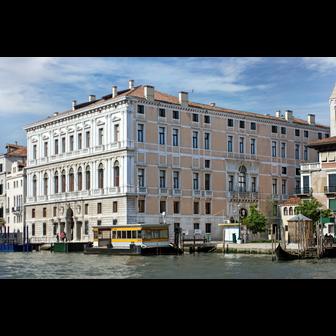} & 
    \includegraphics[width=2cm]{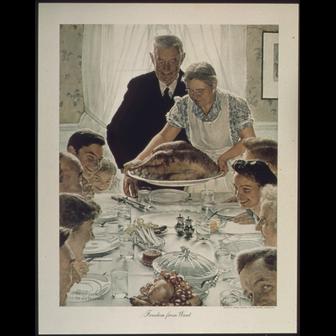} &
    \includegraphics[width=2cm]{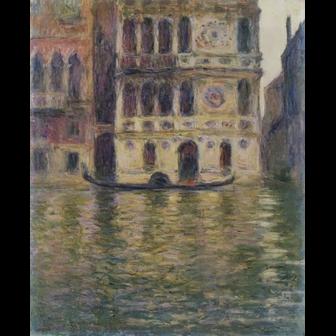} \\
    \cmidrule(lr){2-5}
    \multirow{-4}{*}{Cap-Only (by FLUX)} &
    \includegraphics[width=2cm]{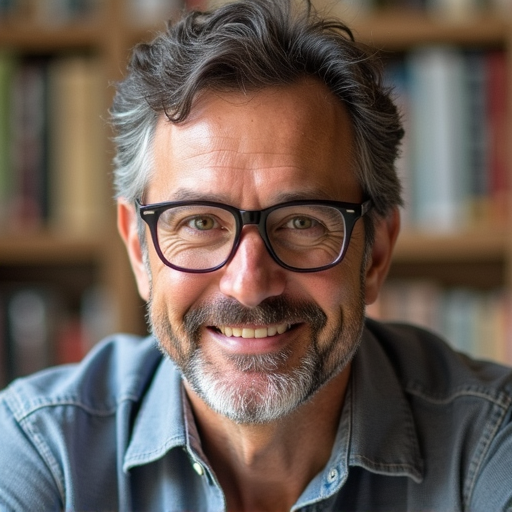} &
    \includegraphics[width=2cm]{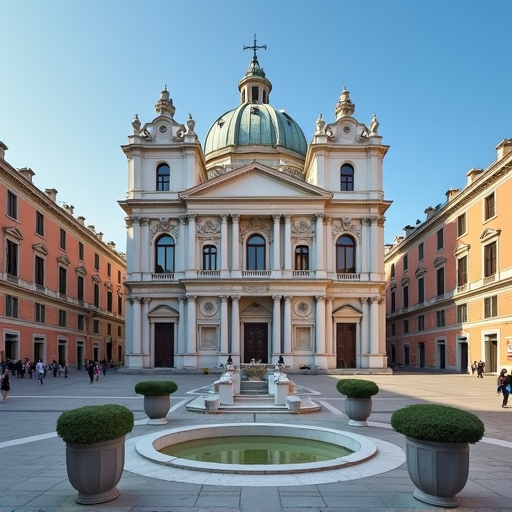} &
    \includegraphics[width=2cm]{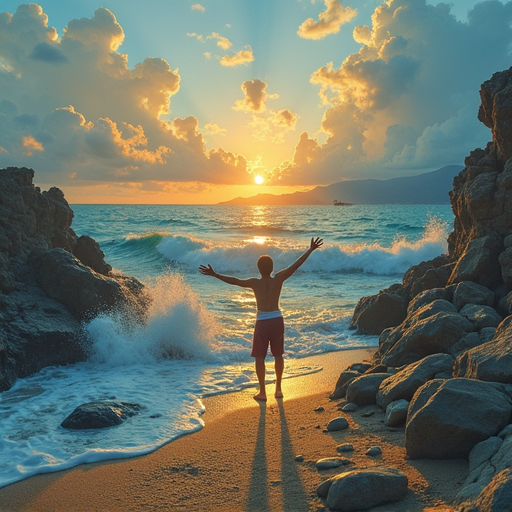} &
    \includegraphics[width=2cm]{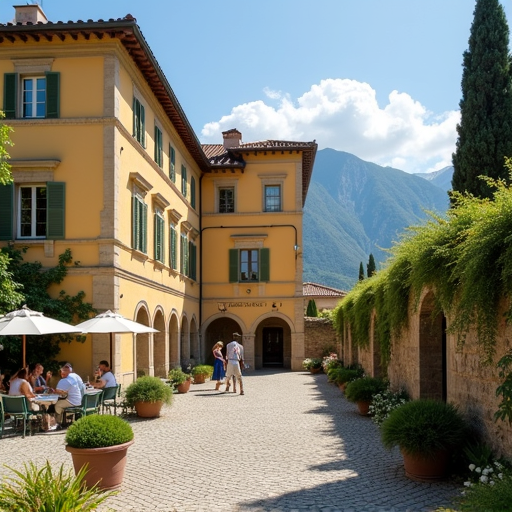} \\
    & 1.03 / 1 / 1/ 1 & 57.23 / 4 / 3 / 4 & 3.03 / 1 / 1 / 1 & 40.23 / 2 / 2 / 3 \\
    \midrule
    \multicolumn{5}{l}{\textbf{Proposed Method (TextTIGER)}} \\
    \midrule
    \multirow{-4}{*}{Dreamlike} & 
    \includegraphics[width=2cm]{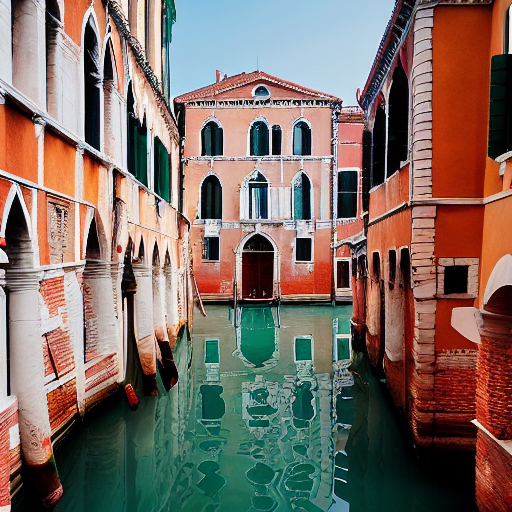} &
    \includegraphics[width=2cm]{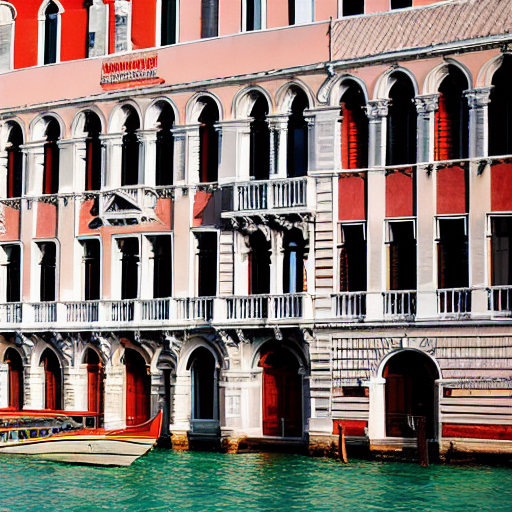} &
    \includegraphics[width=2cm]{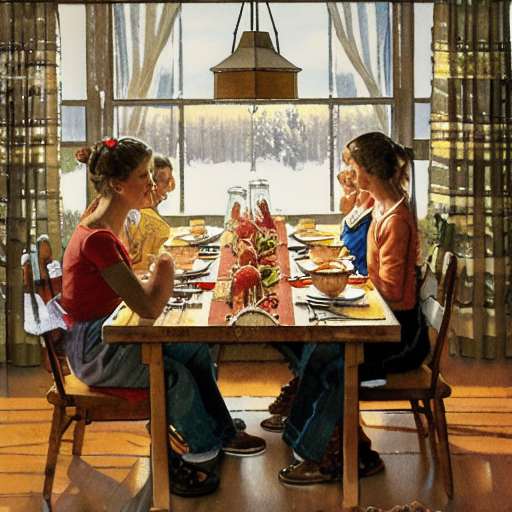} &
    \includegraphics[width=2cm]{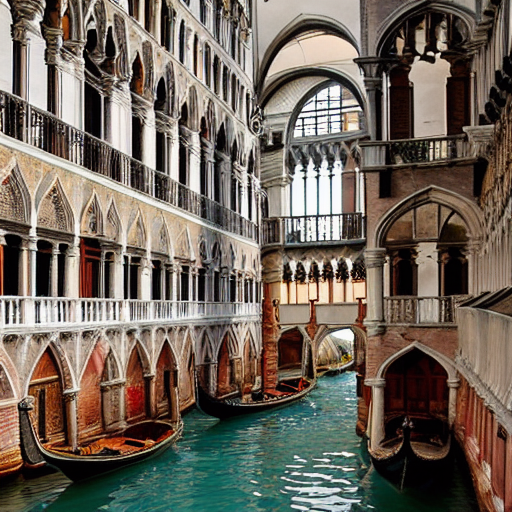} \\
    & 56.23 / 4 / 4 / 4 & 60.45 / 4 / 3 / 3 & 75.93 / 3 / 3 / 3 & 60.23 / 4 / 4 / 4 \\
    \cmidrule(lr){2-5}
    \multirow{-4}{*}{SD3.5} &
    \includegraphics[width=2cm]{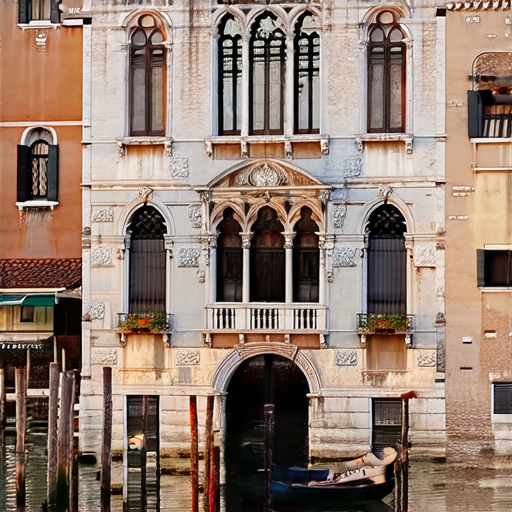} &
    \includegraphics[width=2cm]{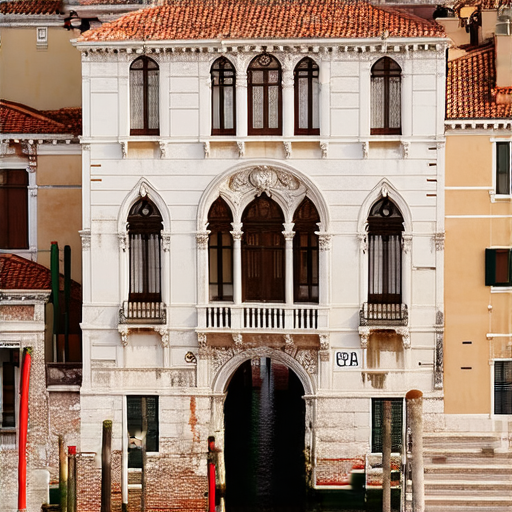} &
    \includegraphics[width=2cm]{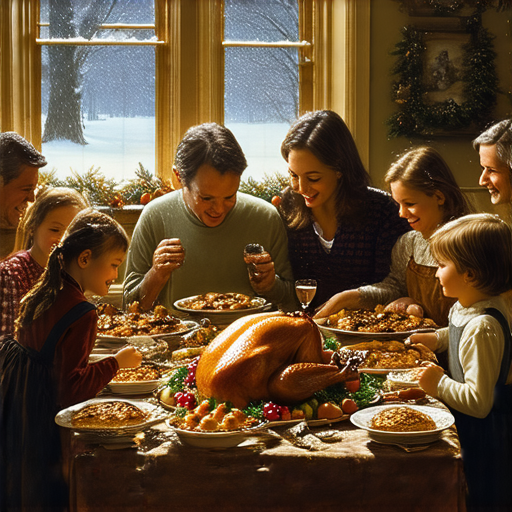} &
    \includegraphics[width=2cm]{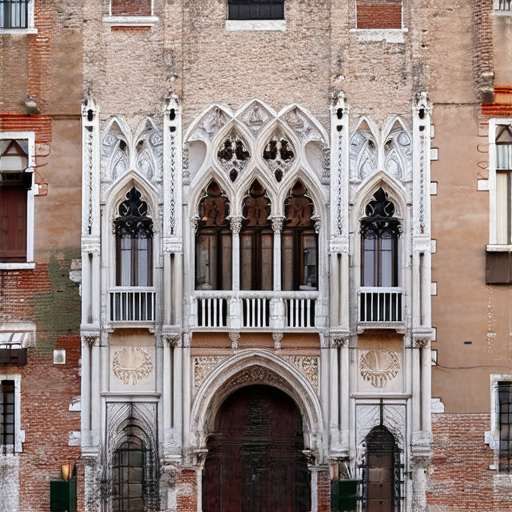} \\
    & 60.06 / 3 / 4 / 4 & 75.75 / 3 / 3 / 4 & 60.23 / 4 / 4 / 4 & 60.22 / 4 / 3 / 5 \\
    \bottomrule
    \end{tabular}
    }
    \label{tab:actual_samepls_summarized_by_qwen3}
\end{table}

\subsection{Qualitative Analysis of Generated Images}
Our work further verifies that the tendency aligns with the actual visual outputs by examining the generated images shown in Tables~\ref{tab:actual_samepls_generated_by_sd3.5} and ~\ref{tab:actual_samepls_summarized_by_qwen3}.
Table~\ref{tab:actual_samepls_generated_by_sd3.5} compares images generated by SD 3.5.
For the Landmarks examples ``Białowieża Forest'' and ``Po-i-Kalyan,'' and the Paintings examples ``Sacred conversation''and ``Solly Madonna,'' \textsc{Cap-Only} produces images that clearly diverge from the reference images.
For instance, in ``Białowieża Forest,'' the model generates a generic forest scene, but it fails to reflect distinctive contextual or symbolic elements.
The DINO score remains as low as 3.12.
In contrast, \textsc{TextTIGER} with Qwen3 summarization raises the score to 85.12, and with Llama3.3 to 90.21.
The generated images visually exhibit a closer match to the atmosphere and composition of the reference forest scene.
A similar pattern appears in ``Po-i-Kalyan.''
\textsc{Cap-Only} achieves only 2.07, whereas \textsc{TextTIGER} improves the score to around 80.
The generated image reproduces characteristic architectural elements such as domes and minarets, and this visual improvement corresponds directly to the large gains in automatic evaluation.
In the Paintings category, ``Sacred conversation'' and ``Solly Madonna'' show abstract or distorted outputs under \textsc{Cap-Only}.
After applying \textsc{TextTIGER}, the images clearly depict structured religious compositions with coherent figure arrangements.
Both DINO and MLLM-based scores (Gemma3, Qwen2.5, Phi4) rise to around 4, matching the visible improvements.
Table~\ref{tab:actual_samepls_summarized_by_qwen3} presents additional examples using Qwen3 summarization.
For ``Ca’ Vendramin Calergi,'' \textsc{Cap-Only} with FLUX yields a near failure case with a score of 1.03.
After applying \textsc{TextTIGER}, the score increases to 56.23 with Dreamlike, 60.06 with SD3.5, and 89.13 with Qwen-Img.
Visually, \textsc{Cap-Only} produces an incorrect portrait-like image, which indicates entity misidentification. 
\textsc{TextTIGER} instead generates a Venetian palace façade that matches the target entity.
The increase in automatic metrics aligns with improved entity recognition.
``Freedom from Want'' exhibits the same pattern.
\textsc{Cap-Only} scores 3.03 and fails to depict the intended scene.
\textsc{TextTIGER} raises the score to the 60--75 range and successfully reconstructs the iconic family dining composition.
DINO and MLLM-based evaluations, many around 4 points, match the visual improvements.

\section{Conclusion}
We addressed the limitations of current text-to-image generation models in handling entity-specific knowledge, which is essential for producing user-intended outputs.
To validate this problem, we introduced a novel dataset that enriches image–caption pairs with entity annotations and detailed descriptions.
Leveraging this dataset, we proposed \textsc{TextTIGER}, a method that augments prompts with external information and uses LLMs to summarize the information, ensuring the inclusion of essential knowledge while keeping the prompt within a length suitable for image generation models.
Our experiments demonstrated that \textsc{TextTIGER} consistently outperforms baseline approaches across both automatic metrics MLLM-as-a-judge.
These results confirm that entity-aware prompt refinement is a promising direction for improving reliability.

\bibliography{main}
\bibliographystyle{tmlr}

\appendix
\section{Future Work}
This study leaves several directions for future work.

First, we acknowledge the absence of human evaluation.
In our work, we demonstrated that our method improves image generation performance using KITTEN, an entity-based evaluation framework shown to have a strong correlation with human judgment in evaluating whether generated images contain the intended entities.
Through this framework, we verified that our approach enhances entity-level controllability in image generation.
However, a direct human evaluation remains an important complementary validation.
Conducting large-scale human studies, particularly for image generation tasks, incurs substantial financial and logistical costs.
Due to budget constraints beyond the scope of this project, we were unable to perform such evaluations. 
Thus, we leave comprehensive human evaluation as an important direction for future work.

Second, regarding the choice of image generation models, we evaluated our method on 5 open models but did not include proprietary models such as GPT-Image.
\footnote{\url{https://developers.openai.com/api/docs/guides/image-generation/}}
While we demonstrated that our approach is effective across different text encoder settings, it is in principle applicable to API-based commercial models as well.
However, since such models require per-generation API costs, conducting large-scale controlled experiments would necessitate substantial additional costs.
For this reason, extending the evaluation to proprietary models is also left for future work.

Finally, further investigation into summarization evaluation is possible.
Our results indicate that improved summarization quality leads to better image generation performance.
Although automatic summarization metrics such as ROUGE~\citep{lin-2004-rouge} could be employed for additional analysis, higher ROUGE scores do not necessarily correspond to prompts that are optimal for image generation.
Because our objective is to improve generation performance rather than maximize textual overlap with reference summaries, we did not include ROUGE-based evaluation in this study.
Exploring the relationship between conventional summarization metrics and downstream image generation quality remains an interesting direction for future research.

\section{Ethical Considerations}
The data used in our study were created using links that were valid as of December 2025.
However, because the images and articles referenced by these links depend on Wikipedia, they may be subject to edits by other users.
Thus, the reproducibility of the data cannot be fully guaranteed.
The code and dataset used in our work will be made publicly available upon acceptance.

\section{Appendix}
\subsection{Detailed Model Settings}
\label{detailed-model-settings}
Table~\ref{tab:detailed_model_names} shows lists of detailed model names used in our experiment.

\begin{table}[t]
    \centering
    \small
    \caption{Detailed models' name and their citations.}
    \begin{tabular}{cccc}
    \toprule
    \textbf{Model} & \textbf{Params.} & \textbf{HuggingFace / OpenAI API Name} & \textbf{Citation} \\
    \midrule
    \rowcolor{red!10}
    \multicolumn{4}{c}{\textbf{Summarization Models}} \\
    \midrule
    Qwen3 & 30B & Qwen/Qwen3-30B-A3B-Instruct-2507 & \citet{yang2025qwen3technicalreport} \\
    Llama 3.3 & 70B & meta-llama/Llama-3.3-70B-Instruct & \citet{grattafiori2024llama3herdmodels} \\
    Qwen2.5 & 72B & Qwen/Qwen2.5-72B-Instruct & \citet{qwen2025qwen25technicalreport} \\
    GPT-4o & -- & gpt-4o-mini-2024-07-18 &  \citet{openai2024gpt4ocard} \\
    GPT-5 & -- & gpt-5-nano-2025-08-07 & \citet{singh2025openaigpt5card} \\
    \midrule
    \rowcolor{yellow!10}
    \multicolumn{4}{c}{\textbf{Image Generation Models}} \\
    \midrule    
    Dreamlike & -- & dreamlike-art/dreamlike-photoreal-2.0 & \citet{dreamlike_photoreal_2} \\
    PixArt & -- & PixArt-alpha/PixArt-XL-2-1024-MS & \citet{chen2023pixartalphafasttrainingdiffusion} \\
    FLUX & 12B & black-forest-labs/FLUX.1-dev & \citet{flux2024} \\
    SD3.5 & -- & stabilityai/stable-diffusion-3.5-large & \citet{esser2024scalingrectifiedflowtransformers} \\ 
    Qwen-Img & -- & Qwen/Qwen-Image & \citet{wu2025qwenimagetechnicalreport} \\
    \midrule
    \rowcolor{green!10}
    \multicolumn{4}{c}{\textbf{MLLM-as-a-judge Models}} \\
    \midrule
    Gemma 3 & 4B & google/gemma-3-4b-it &  \citet{gemmateam2025gemma3technicalreport} \\
    Phi 4 & 6B & microsoft/Phi-4-multimodal-instruct & \citet{microsoft2025phi4minitechnicalreportcompact} \\
    Qwen 2.5-VL & 7B & Qwen/Qwen2.5-VL-7B-Instruct & \citet{bai2025qwen25vltechnicalreport} \\
    \midrule
    \rowcolor{blue!10}
    \multicolumn{4}{c}{\textbf{Retriever Models and Embedding Models}} \\
    \midrule
    BGE & 0.1B & BAAI/bge-base-en-v1.5 & \citet{10.1145/3626772.3657878} \\
    Contriever & 0.1B & facebook/contriever & \citet{lei-etal-2023-unsupervised} \\
    E5 & 0.1B & intfloat/e5-base & \citet{wang2024textembeddingsweaklysupervisedcontrastive} \\
    \midrule
    RoBERTa & 0.1B & FacebookAI/roberta-base & \citet{liu2019robertarobustlyoptimizedbert} \\
    SimCSE & 0.1B & princeton-nlp/sup-simcse-roberta-large & \citet{gao-etal-2021-simcse} \\
    \bottomrule
    \end{tabular}
    \label{tab:detailed_model_names}
\end{table}

\paragraph{Summarization Task}
All experiments were conducted using the Transformers library~\citep{wolf-etal-2020-transformers} with the random seed fixed at 42 for reproducibility.
Qwen2.5 / 3, and Llama3.3 were quantized to 4-bit precision using the bitsandbytes library~\citep{dettmers2022gpt3}.

\paragraph{Image Generation Task}
All experiments were carried out using the diffusers library~\citep{von-platen-etal-2022-diffusers}.
The image resolution was fixed at 512 $\times$ 512 pixels.
The guidance scale was set to 4.5, the number of inference steps to 30, and the random seed to 42.

\begin{wraptable}{r}{0.45\linewidth}
    \vspace{-5mm}
    \small
    \centering
    \caption{Comparison of retriever accuracy (\%).}
    \begin{tabular}{cc}
    \toprule
    \textbf{Retriever} & \textbf{Correct (\%)} \\
    \midrule
    \rowcolor{gray!10}
    \multicolumn{2}{l}{\textbf{Sparse retriever}} \\
    \midrule
    BM25 & \textbf{31.99} \\
    \midrule
    \rowcolor{gray!10}
    \multicolumn{2}{l}{\textbf{Dense retriever}} \\
    \midrule
    BGE & 31.58 \\
    Contriever & 31.36 \\
    E5 & 31.65 \\
    \bottomrule
    \end{tabular}
    \label{tab:retriever-comparison}
    \vspace{-5mm}
\end{wraptable}

\subsection{Difference in Retriever Performance}
\label{app:difference-bw-retriever}
Table~\ref{tab:retriever-comparison} presents the difference in retriever performance.
Preliminary experiments compared both sparse and dense retrievers to determine the most suitable retrieval method for our RAG-based setting.
As shown in Table~\ref{tab:retriever-comparison}, BM25 achieved the highest accuracy (31.99\%), slightly outperforming dense retrievers such as BGE, Contriever, and E5.

Although the performance differences are relatively small, BM25 consistently yielded the best results among the candidates.
Based on this observation, we selected BM25 as the retriever for the experiments reported in the main paper. 

\subsection{Can a RAG Approach Serve as a Substitute?}
The results in Tables~\ref{tab:overall-results-landmarks} and~\ref{tab:overall-results-paintings} show that a naive RAG approach (\textsc{Aug-Only}, \textsc{BM25}) cannot sufficiently substitute our method.

For example, in the Landmarks category, CLIPScore-T for Dreamlike drops from 23.978 under \textsc{Cap-Only} to 20.939 with \textsc{Aug-Only} and 20.743 with \textsc{BM25}.
Rather than improving performance, \textsc{RAG} degrades it.
We observe similar trends for PixArt, FLUX, and Qwen-Img, where RAG fails to consistently outperform \textsc{Cap-Only}.
In the Paintings category, RAG occasionally produces small improvements for certain metrics.
However, it does not match the consistent and substantial gains achieved by \textsc{TextTIGER}, suggesting that simply retrieving and appending external documents increases noise and redundancy, which prevents the model from effectively leveraging critical entity information.
Image generation models face constraints in input token length and attention allocation.
Thus, directly injecting unstructured retrieval outputs does not necessarily work well.

Although RAG provides a general framework for leveraging external knowledge, it does not replace our entity-focused summarization and structured knowledge injection.
Challenges such as retrieval accuracy, summarization quality, and noise reduction remain unresolved, leaving further performance improvements through RAG-style approaches for future work.

\subsection{Does the Choice of Text Encoder Affect Performance?}
The results in Tables~\ref{tab:overall-results-landmarks} and ~\ref{tab:overall-results-paintings} indicate that the choice of text encoder influences baseline performance to some extent.
However, the effect of \textsc{TextTIGER} consistently appears across encoder types.
First, we examine Dreamlike and SD3.5, which use CLIP as the text encoder.
When we move from \textsc{Cap-Only} to \textsc{TextTIGER}, all evaluation metrics improve substantially.
For instance, in Landmarks, CLIPScore-T for Dreamlike increases from 23.978 to a maximum of 25.296, and for SD3.5 from 23.882 to 25.475.
We observe similar gains in CLIPScore-I, DINOScore, and PickScore, indicating that external knowledge injection strongly enhances entity understanding even for CLIP-based encoders.

Next, we analyze PixArt, which uses the T5 encoder.
Under \textsc{Cap-Only}, their scores remain comparable to or slightly lower than CLIP-based models.
However, applying \textsc{TextTIGER} consistently improves performance.
In Landmarks, CLIPScore-T for PixArt rises from 19.106 to 23.244, and DINOScore and PickScore show similar improvements.
Our result observes the same pattern in Paintings, indicating that entity-specific knowledge augmentation benefits T5-based encoders as well.
Therefore, the primary driver of performance gains lies in external knowledge injection rather than in any specific encoder architecture.

\begin{table*}[h]
\centering
\caption{Hits@K results across categories. Best results per column are in bold.}
\resizebox{\linewidth}{!}{
\begin{tabular}{llcccccccc}
\toprule
 \multirow{2.5}{*}{\textbf{Model}} & \multirow{2.5}{*}{\textbf{Encoder}} & \multicolumn{4}{c}{\textbf{Landmarks}} & \multicolumn{4}{c}{\textbf{Paintings}} \\
\cmidrule(lr){3-6}\cmidrule(lr){7-10}
& & \textbf{Hits@1} & \textbf{Hits@5} & \textbf{Hits@10} & \textbf{Hits@100} & \textbf{Hits@1} & \textbf{Hits@5} & \textbf{Hits@10} & \textbf{Hits@100} \\
\midrule
Dreamlike2.0 & CLIP & 0.000 & 0.002 & 0.004 & 0.036 & 0.000 & 0.002 & 0.004 & 0.045 \\
SD3.5 & CLIP-G & 0.007 & 0.016 & 0.025 & 0.111 & 0.002 & 0.014 & 0.022 & 0.084 \\
SD3.5 & CLIP-L & 0.190 & 0.367 & 0.451 & 0.765 & 0.060 & 0.131 & 0.175 & 0.369 \\
SD3.5 & T5 & 0.012 & 0.021 & 0.032 & 0.167 & 0.003 & 0.012 & 0.022 & 0.104 \\
PixArt & T5 & 0.004 & 0.009 & 0.012 & 0.068 & 0.011 & 0.019 & 0.023 & 0.095 \\
FLUX & CLIP & 0.000 & 0.002 & 0.004 & 0.036 & 0.000 & 0.002 & 0.004 & 0.045 \\
FLUX & T5 & 0.005 & 0.008 & 0.012 & 0.066 & 0.011 & 0.022 & 0.029 & 0.100 \\
Qwen-Img & Qwen (LLM) & 0.009 & 0.024 & 0.034 & 0.130 & 0.004 & 0.009 & 0.017 & 0.103 \\
\hdashline
RoBERTa & BERT & 0.029 & 0.064 & 0.089 & 0.238 & 0.025 & 0.042 & 0.059 & 0.178 \\
SimCSE & BERT & \textbf{0.306} & \textbf{0.489} & \textbf{0.563} & \textbf{0.809} & \textbf{0.181} & \textbf{0.314} & \textbf{0.362} & \textbf{0.579} \\
\bottomrule
\end{tabular}
}
\label{tab:hits-results}
\end{table*}

\clearpage
\subsection{Examples of Our Created Dataset}
\label{detailed-dataset-settngs}
Tables~\ref{tab:dataset-example-of-paintings} and ~\ref{tab:dataset-example-of-landmarks} show examples of our created dataset.

\begin{table}[h]
    \centering
    \caption{An example of the created dataset in Paintings category.}
    \resizebox{\linewidth}{!}{
    \begin{tabular}{cp{0.6\linewidth}c}
    \toprule
    \textbf{Entity} & \textbf{Description} & \textbf{Ref. Image} \\
    \midrule
Queen Victoria & Victoria was \underline{Queen of the United Kingdom} of \underline{Great Britain} and Ireland from 20 June 1837 until her death in 1901. Her reign of 63 years and 216 days, which was longer than those of any of her predecessors, constituted the \underline{Victorian era}, a period of industrial, political, scientific, and military change within the United Kingdom marked by a great expansion of the \underline{British Empire}. In 1876, the \underline{British parliament} voted to grant her the additional title of \underline{Empress of India}. & \multirow{40}{*}{\includegraphics[width=2cm]{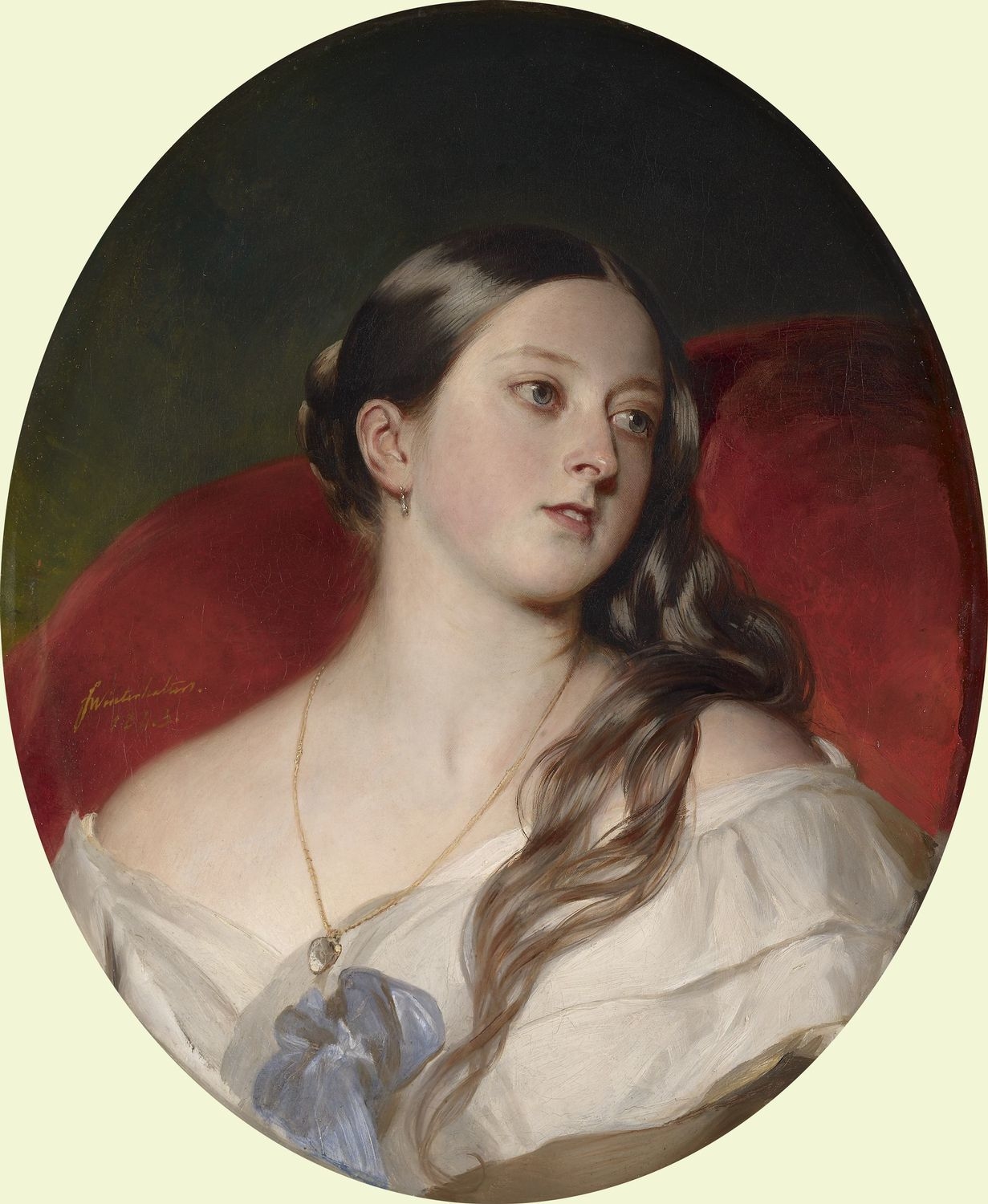}} \\
     \cdashline{1-2}
British Empire & The British Empire comprised the dominions, colonies, protectorates, mandates, and other territories ruled or administered by the United Kingdom and its predecessor states. It began with the overseas possessions and trading posts established by England in the late 16th and early 17th centuries, and colonisation attempts by Scotland during the 17th century. At its height in the 19th and early 20th centuries, it became the largest empire in history and, for a century, was the foremost global power. By 1913, the British Empire held sway over 412 million people, 23 percent of the world population at the time, and by 1920, it covered 35.5 million km2 (13.7 million sq mi), 24 per cent of the Earth's total land area. As a result, its constitutional, legal, linguistic, and cultural legacy is widespread. At the peak of its power, it was described as "the empire on which the sun never sets", as the sun was always shining on at least one of its territories. \\
     \cdashline{1-2}
Great Britain & Great Britain is an island in the North Atlantic Ocean off the north-west coast of continental Europe, consisting of the countries England, Scotland and Wales. With an area of 209,331 km2 (80,823 sq mi), it is the largest of the British Isles, the largest European island, and the ninth-largest island in the world. It is dominated by a maritime climate with narrow temperature differences between seasons. The island of Ireland, with an area 40 per cent that of Great Britain, is to the west – these islands, along with over 1,000 smaller surrounding islands and named substantial rocks, comprise the British Isles archipelago. \\
     \cdashline{1-2}
\multirow{2}{*}{\shortstack{United Kingdom of \\ Great Britain and Ireland}} & The United Kingdom of Great Britain and Ireland was established by the Acts of Union in 1801 that united the Kingdom of Great Britain and the Kingdom of Ireland into one sovereign state. It continued in this form until 1927, when it evolved into the United Kingdom of Great Britain and Northern Ireland, after the Irish Free State gained a degree of independence in 1922. \\
     \cdashline{1-2}
Victorian era & In the history of the United Kingdom and the British Empire, the Victorian era was the reign of Queen Victoria, from 20 June 1837 until her death on 22 January 1901, although slightly different definitions are sometimes used. The era followed the Georgian era and preceded the Edwardian era, and its later half overlaps with the first part of the Belle Époque era of continental Europe. \\
    \bottomrule
    \end{tabular}
    }
    \label{tab:dataset-example-of-paintings}
\end{table}

\begin{table}[t]
    \centering
    \caption{An example of the created dataset in Landmarks category.}
    \resizebox{\linewidth}{!}{
    \begin{tabular}{cp{0.6\linewidth}c}
    \toprule
    \textbf{Entity} & \textbf{Description} & \textbf{Ref. Image} \\
    \midrule
Taj Mahal & The Taj Mahal is an ivory-white marble mausoleum on the right bank of the river \underline{Yamuna} in \underline{Agra}, \underline{Uttar Pradesh}, \underline{India}. It was commissioned in 1631 by the fifth Mughal emperor, \underline{Shah Jahan}, to house the tomb of his beloved wife, \underline{Mumtaz Mahal}; it also houses the tomb of \underline{Shah Jahan} himself. The tomb is the centrepiece of a 17-hectare (42-acre) complex, which includes a mosque and a guest house, and is set in formal gardens bounded on three sides by a crenellated wall. & \multirow{40}{*}{\includegraphics[width=2cm]{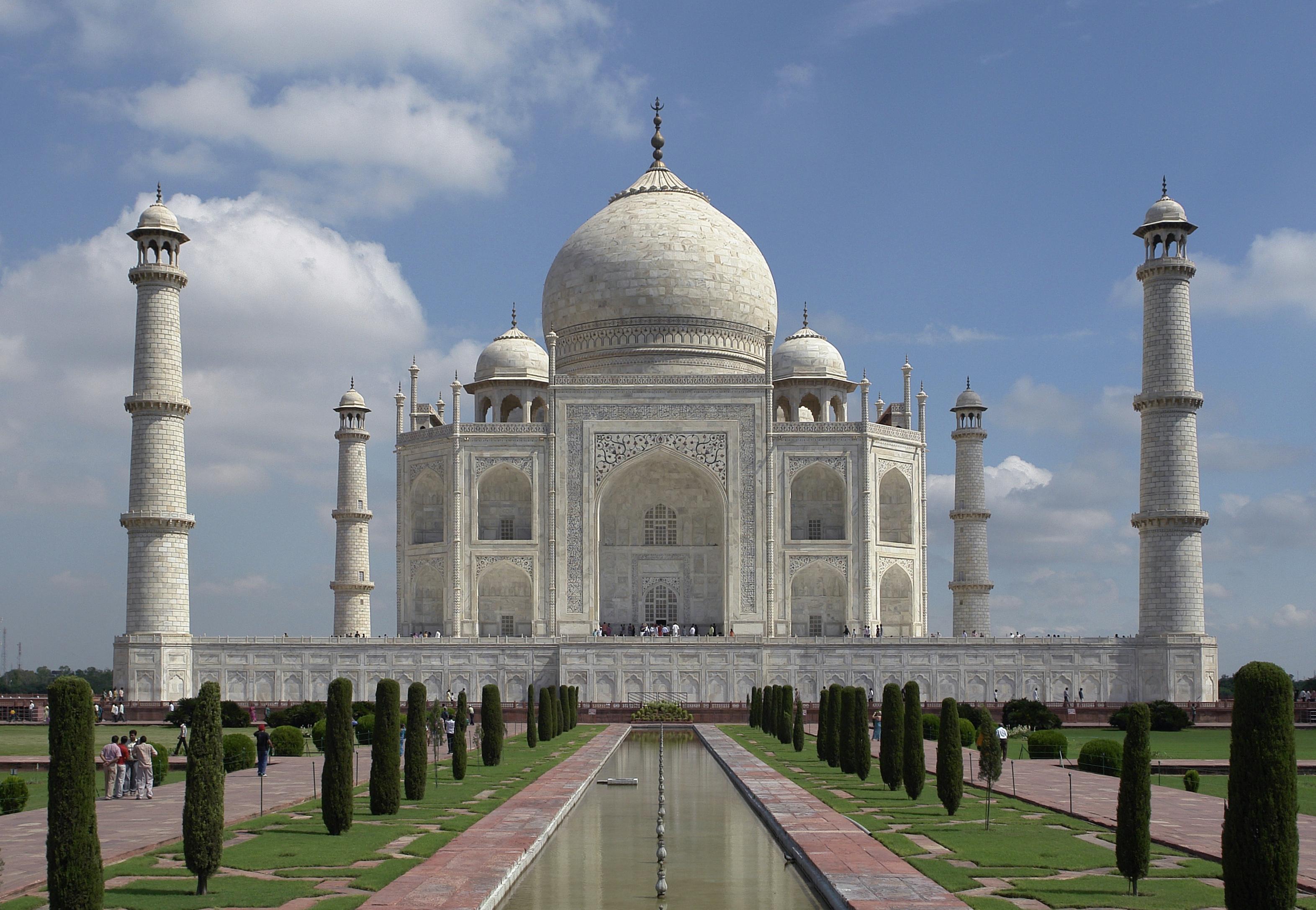}} \\
     \cdashline{1-2}
Agra & Agra is a city on the banks of the Yamuna river in the Indian state of Uttar Pradesh, about 230 kilometres (140 mi) south-east of the national capital Delhi and 330 km west of the state capital Lucknow. It is also the part of Braj region. With a population of roughly 1.6 million, Agra is the fourth-most populous city in Uttar Pradesh and twenty-third most populous city in India. \\
     \cdashline{1-2}
India & India, officially the Republic of India, is a country in South Asia. It is the seventh-largest country by area; the most populous country since 2023; and, since its independence in 1947, the world's most populous democracy. Bounded by the Indian Ocean on the south, the Arabian Sea on the southwest, and the Bay of Bengal on the southeast, it shares land borders with Pakistan to the west; China, Nepal, and Bhutan to the north; and Bangladesh and Myanmar to the east. In the Indian Ocean, India is near Sri Lanka and the Maldives; its Andaman and Nicobar Islands share a maritime border with Myanmar, Thailand, and Indonesia. \\
     \cdashline{1-2}
Marble & Marble is a metamorphic rock consisting of carbonate minerals (most commonly calcite (CaCO3) or dolomite (CaMg(CO3)2) that have recrystallized under the influence of heat and pressure. It has a crystalline texture, and is typically not foliated (layered), although there are exceptions. \\
     \cdashline{1-2}
Mausoleum & A mausoleum is an external free-standing building or standalone structure constructed as a monument enclosing the burial chamber of a deceased person or people. A mausoleum without the person's remains is called a cenotaph. A mausoleum may be considered a type of tomb, or the tomb may be considered to be within the mausoleum. \\
Mosque & A mosque, also called a masjid, is a place of worship for Muslims. The term usually refers to a covered building, but can be any place where Islamic prayers are performed; such as an outdoor courtyard. \\
     \cdashline{1-2}
Mumtaz Mahal & Mumtaz Mahal was the empress consort of Mughal Empire from 1628 to 1631 as the chief consort of the fifth Mughal emperor, Shah Jahan. The Taj Mahal in Agra, often cited as one of the Wonders of the World, was commissioned by her husband to act as her tomb. \\
    \bottomrule
    \end{tabular}
    }
    \label{tab:dataset-example-of-landmarks}
\end{table}

\clearpage
\section{Prompts}
We list the prompts used during experiments below.

\subsection{Prompts for Summarization Task}
\label{prompts-for-summarization}

\begin{tcolorbox}[title=Prompt for Summarization (\textsc{TextTIGER}), boxrule=1pt, colback=white, coltitle=black, colframe=blue!50, colbacktitle=blue!60]
\scriptsize
You must generate **ONLY ONE THING**: \\
a single English prompt for an image-generation model. \\ 
Do NOT output explanations, comments, apologies, thoughts, or any other text. \\ 

\#\#\# HARD OUTPUT FORMAT CONSTRAINT \\
You MUST output **ONLY** the following block EXACTLY in this format: \\

<SummaryStart> \\
English prompt for the image generator, within 70 tokens, nothing else \\
<SummaryEnd> \\

- No additional text before or after the tags. \\
- No reasoning steps. \\
- No markdown. \\
- No prefaces or suffixes. \\
- No self-talk. \\
- No comments. \\ 
- No variable placeholders. \\

\#\#\# TASK (STRICT) \\
Create the **optimal** English prompt for generating an iconic image of **\texttt{\{title\}}**. \\

Use only information logically inferable from the summary below. \\
Assume the image-generation model does NOT know what ``\texttt{\{title\}}'' is. \\

\#\#\# REQUIREMENTS \\
- Length: **≤ 70 tokens** \\
- Include **all concrete visual details** required for correct generation: \\
  - environment (sea, mountains, city, interior, etc.) \\
  - physical structure, shapes \\
  - materials, colors \\
  - atmosphere, lighting \\
  - perspective or composition \\
  - style (only if described or inferable) \\
  - measurements (height/width) if included in the summary \\
- Do NOT include: \\
  - citations \\
  - mentions of “summary,” “tokens,” or the instructions \\
  - analysis or meta text \\
  - any text outside <SummaryStart> … <SummaryEnd> \\
- Generate \\
  - ONLY prompt \\
  - ONLY in English \\
  - ONLY 1 sentence \\

\#\#\# REFERENCE SUMMARY \\
Below is the summary of \texttt{\{abstract\_tokens\}} tokens. \\
Use ONLY the information contained in it. \\

--- SUMMARY BELOW --- \\
\texttt{\{abstract\}} \\
--- END SUMMARY ---  \\

Now output ONLY the required block: \\

<SummaryStart> \\
... \\
<SummaryEnd> \\

<SummaryStart> \\
\end{tcolorbox}

\begin{tcolorbox}[title=Prompt for Summarization (\textsc{TextTIGER w/o Len}), boxrule=1pt, colback=white, coltitle=black, colframe=green!80, colbacktitle=green!80]
\scriptsize
You must generate **ONLY ONE THING**: \\
a single English prompt for an image-generation model. \\
Do NOT output explanations, comments, apologies, thoughts, or any other text. \\

\#\#\# HARD OUTPUT FORMAT CONSTRAINT \\
You MUST output **ONLY** the following block EXACTLY in this format: \\

<SummaryStart> \\
English prompt for the image generator, nothing else \\
<SummaryEnd>  \\

- No additional text before or after the tags. \\
- No reasoning steps. \\
- No markdown. \\
- No prefaces or suffixes. \\
- No self-talk. \\
- No comments. \\
- No variable placeholders. \\

\#\#\# TASK (STRICT) \\
Create the **optimal** English prompt for generating an iconic image of **\texttt{\{title\}}**. \\

Use only information logically inferable from the summary below. \\
Assume the image-generation model does NOT know what ``\texttt{\{title\}}'' is. \\

\#\#\# REQUIREMENTS \\
- Include **all concrete visual details** required for correct generation: \\
  - environment (sea, mountains, city, interior, etc.) \\
  - physical structure, shapes \\
  - materials, colors \\
  - atmosphere, lighting \\
  - perspective or composition \\
  - style (only if described or inferable) \\ 
  - measurements (height/width) if included in the summary \\
- Do NOT include: \\
  - citations \\
  - mentions of ``summary''  or the instructions \\
  - analysis or meta text \\
  - any text outside <SummaryStart> … <SummaryEnd> \\
- Generate \\
  - ONLY prompt \\
  - ONLY in English \\

--- SUMMARY BELOW --- \\
\texttt{\{abstract\}} \\
--- END SUMMARY --- \\

Now output ONLY the required block: \\

<SummaryStart> \\
... \\
<SummaryEnd> \\

<SummaryStart> \\
\end{tcolorbox}

\subsection{Prompts for MLLM-as-a-judge (KITTEN)}
\label{prompts-for-kitten}

\begin{tcolorbox}[title=Prompt for Entity Alignment, boxrule=1pt, colback=white, coltitle=black, colframe=yellow!90, colbacktitle=yellow!90]
\scriptsize
In this task, you will be provided with a Prompt, Reference Images, and a Generated Image. Your
job is to assess the factual accuracy of the generated image with respect to the prompt and the
reference images. The goal is to ensure that the entity described in the prompt is factually correct and
accurately represented. \\

First, evaluate how faithfully the generated image represents the reference entity. \\
Consider whether the key features and overall appearance of the reference entity are accurately depicted.\\
Question 1: How faithfully does the generated image represent the entity mentioned in the prompt? \\

Candidate Answers:\\
1 (Not faithful at all): The generated image does not represent the reference entity at all. \\
There are no discernible visual similarities to the reference entity. \\
2 (Barely faithful): The generated image faintly represents the reference entity, with significant effort needed to see any resemblance. \\
Minor visual elements may be present, but crucial features or characteristics are missing or significantly misrepresented. \\
3 (Somewhat faithful): The generated image somewhat represents the reference entity, but it's not prominent. 
There is a clear visual connection in terms of composition, style, or some key elements, but there are noticeable differences, omissions, or misinterpretations. \\
4 (Mostly faithful): The generated image mostly represents the reference entity and clearly presents it. 
The generated image draws strong visual inspiration with a strong connection in terms of overall composition, style, key elements, and/or subject matter, despite some variations in details. \\
5 (Completely faithful): The generated image fully represents the reference entity accurately.  \\
It captures all key elements, composition, and style in a way that is almost identical to the reference entity. \\

Answer in the exact format below: \\
Question 1: \\
Answer: [1-5] \\
Reason: [Provide a clear explanation for your answer] \\
\end{tcolorbox}

\begin{tcolorbox}[title=Prompt for Prompt Alignment, boxrule=1pt, colback=white, coltitle=black, colframe=red!60, colbacktitle=red!60]
\scriptsize

In this task, you will be provided with a Prompt and a Generated Image. \\\\

Evaluate how well the generated image captures all aspects described in the prompt. \\
Focus on background elements, contextual details, materials, styles, and other visual features. \\

Question 2: How well does the generated image depict the details described in the prompt? \\

Candidate Answers: \\
1 (Not at all): None of the described elements are present in the image. \\
2 (Slightly): A few minor elements are present, but most are missing or inaccurate. \\
3 (Moderately): Some elements are present and somewhat accurate, but others are missing or misrepresented. \\
4 (Mostly): Most of the described elements are clearly and accurately depicted.\\
5 (Completely): All relevant aspects of the prompt are thoroughly and accurately represented. \\

Answer in the exact format below: \\
Question 2:\\
Answer: [1-5] \\
Reason: [Provide a clear explanation for your answer] \\
\end{tcolorbox}

\end{document}